\documentclass[nohyperref]{article}

\usepackage{microtype}
\usepackage{graphicx}
\graphicspath{{images/}}
\usepackage{hhline}
\usepackage{subfigure}
\usepackage{booktabs} 
\newcommand\Tstrut{\rule{0pt}{2.6ex}}         
\newcommand\Bstrut{\rule[-0.9ex]{0pt}{0pt}} 

\usepackage{hyperref}

\usepackage[accepted]{icml2023}

\usepackage{amsmath}
\usepackage{amssymb}
\usepackage{mathtools}
\usepackage{amsthm}

\usepackage[capitalize,noabbrev]{cleveref}

\theoremstyle{plain}

\theoremstyle{definition}

\theoremstyle{remark}

\usepackage[textsize=tiny]{todonotes}

\icmltitlerunning{Reliable Measures of Spread in High Dimensional Latent Spaces}

\begin{document}

\twocolumn[
\icmltitle{Reliable Measures of Spread in High Dimensional Latent Spaces}



\icmlsetsymbol{equal}{*}

\begin{icmlauthorlist}
\icmlauthor{Anna C. Marbut}{anna_affil}
\icmlauthor{Katy McKinney-Bock}{katy_affil}
\icmlauthor{Travis J. Wheeler}{travis_affil}
\end{icmlauthorlist}

\icmlaffiliation{anna_affil}{Department of Interdisciplinary Studies, University of Montana, Missoula, MT, USA}
\icmlaffiliation{katy_affil}{Appling LLC, Portland, OR, USA}
\icmlaffiliation{travis_affil}{Department of Pharmacy Practice \& Science, University of Arizona, Tucson, AZ, USA}

\icmlcorrespondingauthor{Anna C. Marbut}{anna.marbut@umontana.edu}
\icmlcorrespondingauthor{Travis J. Wheeler}{twheeler@arizona.edu}

\icmlkeywords{Machine Learning, ICML, Representation Learning, Latent Geometry}

\vskip 0.3in
]

\printAffiliationsAndNotice{}

\begin{abstract}
Understanding geometric properties of the latent spaces of natural language processing models allows the manipulation of these properties for improved performance on downstream tasks. One such property is the amount of data spread in a model's latent space, or how fully the available latent space is being used. We demonstrate that the commonly used measures of data spread, average cosine similarity and a partition function min/max ratio I(V), do not provide reliable metrics to compare the use of latent space across data distributions. We propose and examine six alternative measures of data spread, all of which improve over these current metrics when applied to seven synthetic data distributions. Of our proposed measures, we recommend one principal component-based measure and one entropy-based measure that provide reliable, relative measures of spread and can be used to compare models of different sizes and dimensionalities.

\end{abstract}

\section{Introduction}
The product of many neural network models is a representation of the data input in a high dimensional latent space. The distribution of data in this latent space is often used in the application of the learned model through data clustering for classification, measuring distance between data points to quantify similarity, sampling to generate synthetic data elements, or any number of other downstream tasks. For this reason, understanding and manipulating the geometric properties of models' latent spaces is an area of active research.

One such geometric property is a quantification of how evenly the data is distributed in its latent representation. It has been shown that many common neural network architectures produce highly anisotropic latent spaces \cite{Ethayarajh2019-fl,Liu2018-qb, Mimno2017-mj}, and recent research has demonstrated improved performance on benchmarking tasks using various methods for enforcing more complete use of a model’s latent space \cite{Bihani2021-pc, Kaneko2020-fq, Liang2021-aq, Mu2017-ck}.

The majority of existing work quantifies the spread of data in a latent space with two measures of isotropy, average cosine similarity and a measure introduced by \citet{Mu2017-ck} based on a ratio of principal component loadings. We present seven synthetic data distributions and show that these two measures do not behave as would be expected of a reliable measure of relative data spread. 

We examine six alternative ways to quantify data spread and compare the performance of these measures on the same example distributions. We consider two principal component measures, two ratios of differential entropy approximations, and two measures of relative entropy (Kullback-Leibler Divergence).
We show that all proposed measures behave more intuitively on our evaluation distributions than the two commonly used measures. In particular, we find that our proposed Eigenvalue Early Enrichment score and Vasicek Ratio MSE score most closely mirror our expectations across these example distributions and various numbers of dimensions.
Finally, we investigate the behavior of the best-performing of the alternative data spread metrics on real (not simulated) latent spaces 
produced by a pre-trained Word2Vec model.

\section{Related Work}
\label{relwork}
In a natural language processing (NLP) context, the expressiveness of a model can be directly tied to the dimensionality of its trained word embeddings, with model expressiveness increasing with the number of available dimensions up to the point of severe overfitting \cite{yin2018dimensionality}. However, this increased expressiveness as a function of latent space dimensionality depends on the model effectively using the space across all of these dimensions.

This assumption of an evenly used latent space is not guaranteed, as addressed by a growing body of work that assesses and manipulates the geometry of latent spaces in NLP models. It has been demonstrated that static word embedding models do not use their latent spaces evenly, with \citet{Mimno2017-mj} finding a large positive inner product between individual word vectors and a global mean in Word2Vec, and \citet{Mu2017-ck} finding non-zero global mean vectors and non-uniform distribution of variance across dimensions for several common models. Considering contextual NLP models, \citet{Ethayarajh2019-fl} find a non-zero average cosine similarity between word vectors throughout all layers of ELMo, BERT, and GPT-2.

Several explanations for this uneven use of the latent space explore the relationship between word representations and word frequency. \citet{Gao2019-zg} introduce the concept of ``representation degeneration'' in which common words are handled differently than rare words during training, and word frequency has also been shown to be directly correlated with word vector magnitude \cite{Kobayashi2020-xa} and distance traveled during training \cite{Gong2018-ko}. Similarly, \citet{Mu2017-ck} find that the top two PCA components of static language model latent spaces are largely dedicated to expressing word frequency information.

Attempts to ``correct'' this incomplete use of the latent space have resulted in improvement on common NLP benchmarking tasks. Post-training adjustments include subtracting the global mean vector and removing highly explanatory principal components \cite{Mu2017-ck, Liang2021-aq, Rajaee2021-lw}, using an autoencoder framework toward a similar goal \cite{Kaneko2020-fq}, and learning a transformation into a more uniformly filled non-Euclidean space \cite{Frenzel2019-ec}. Adjustments made during training include minimizing an additional loss function \cite{Liu2018-qb,Gao2019-zg,Wang2019-jn} and using an isotropic Gaussian prior in a VAE \cite{Zhang2022-qm} or during batch normalization \cite{Zhou2020-td}.

Beyond improving benchmark performance, \citet{Liao2020-ll} remove top principal components during an iterative quantization process and \citet{Sablayrolles2018-zh} use a nearest-neighbor entropy approximation (Kozachenko-Leonenko entropy) to enforce uniform spread before quantizing data. Both of these methods find less data loss after compression than quantization methods that do not focus on using the latent space more completely.

\section{Data Spread in High Dimensions}\label{spread}

A theory of maximizing the expressiveness of a model by maximally using the latent space of that model requires that we first define what it means to "maximally use the latent space". While this task may seem intuitive, translating these intuitions into a well-defined, ideal latent space distribution, particularly in high dimensions, presents several complications. Here we explore existing concepts and distributions that can be used to help define this ideal space.

\subsection{Data Spread vs. Isotropy}\label{spread_def}

One concept that comes up in much of the literature in Section \ref{relwork} is that of \emph{isotropy}. The definition of isotropy varies between scientific and mathematical fields, but broadly it is defined as ``[i]dentical in all directions; invariant with respect to direction'' \cite{dict:isotropy}. In the context of data representations, this means that the distribution of data representations would be the same in any direction from the origin. \citet{Rudelson1999-cv} finds that a probability distribution is in isotropic position if its covariance matrix is the identity, and \citet{Zhou2020-td} define a latent space as isotropic if all dimensions have the same variance and are uncorrelated.

Working from these definitions, a measure of isotropy can tell us whether our data are distributed similarly in all available directions, but it will say nothing about what the distribution is in any given direction. For example, data distributed on the shell of a hypersphere are equally isotropic to data distributed in a multivariate normal distribution. Intuitively we'd expect those two data distributions to perform differently on a measure of how fully the available space is being used.

Here we have introduced the term \emph{isotropic} as meaning that a distribution appears to be the same in all directions. In this manuscript, we will use the term \emph{spread} to expand on this notion; a distribution shows more complete \emph{spread} if it evenly spreads points in all directions (isotropy) as well as along each axis/direction. 

\subsection{Reference Distributions}\label{spread_dist}
 By definition, the uniform distribution will maximally fill any shape in any number of dimensions, making it an obvious choice for our ideally-filled latent space. However, while data in a uniform distribution will maximally fill a hypersphere of any radius, the probability abruptly drops to zero outside of that radius. This may not make sense for data representations in a continuous latent space. 

Alternatively, we might consider a multivariate normal distribution ideal for maximally filling a latent space. Like a uniformly filled hypersphere, data in a multivariate normal distribution will be isotropic, but with probability instead approaching zero gradually as distance from the origin increases. Although using a multivariate normal distribution will result in having a higher density of points around the origin than we would find in referencing a uniform distribution, this gradual decrease in probability (and point density) generally seems a realistic and desirable trait for data representations in a latent space.

\section{Models for Comparison}\label{sec:sample_models}

While we could have used the actual output of popular NLP models to compare different ways of measuring data spread in high dimensional space, we don't necessarily have strong intuition about how fully the data representations in these models fill the latent space or how they should compare to each other on a sufficient measure of data spread. Therefore we developed a collection of seven structured distributions, each with $d=$~$\{2, 10, 50, 100\}$ dimensions and $250d$ data points, and used these as intuitive benchmarks on our proposed measures of spread.  

For spherical distributions, we created a \emph{Shell}, a \emph{Nested Shell}, a uniformly filled \emph{Sphere}, and a \emph{Cone}. For cluster-based distributions, we created clusters that are symmetric across the origin and identical in size (\emph{Symmetric Clusters}), asymmetric clusters of identical size (\emph{Shifted Clusters}), and symmetric clusters of unequal size (\emph{Uneven Clusters}). Details of the characteristics of these distributions can be found in Appendix~\ref{app:dists}.

\begin{figure}[!b]
\begin{center}
\centerline{\includegraphics[width = 0.9\columnwidth]{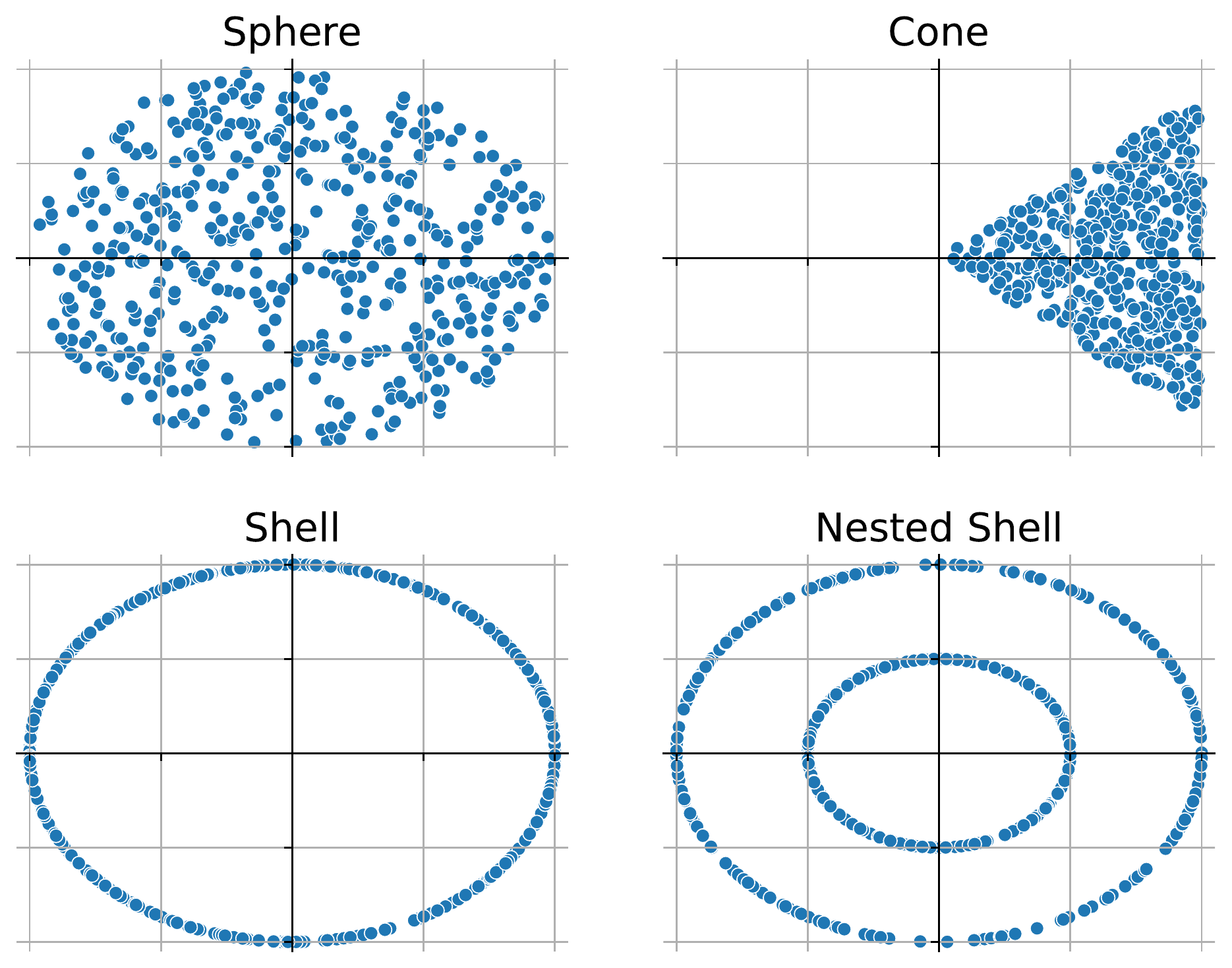}}
\par\vspace{0.1in}
\centerline{\includegraphics[width = 0.9\columnwidth]{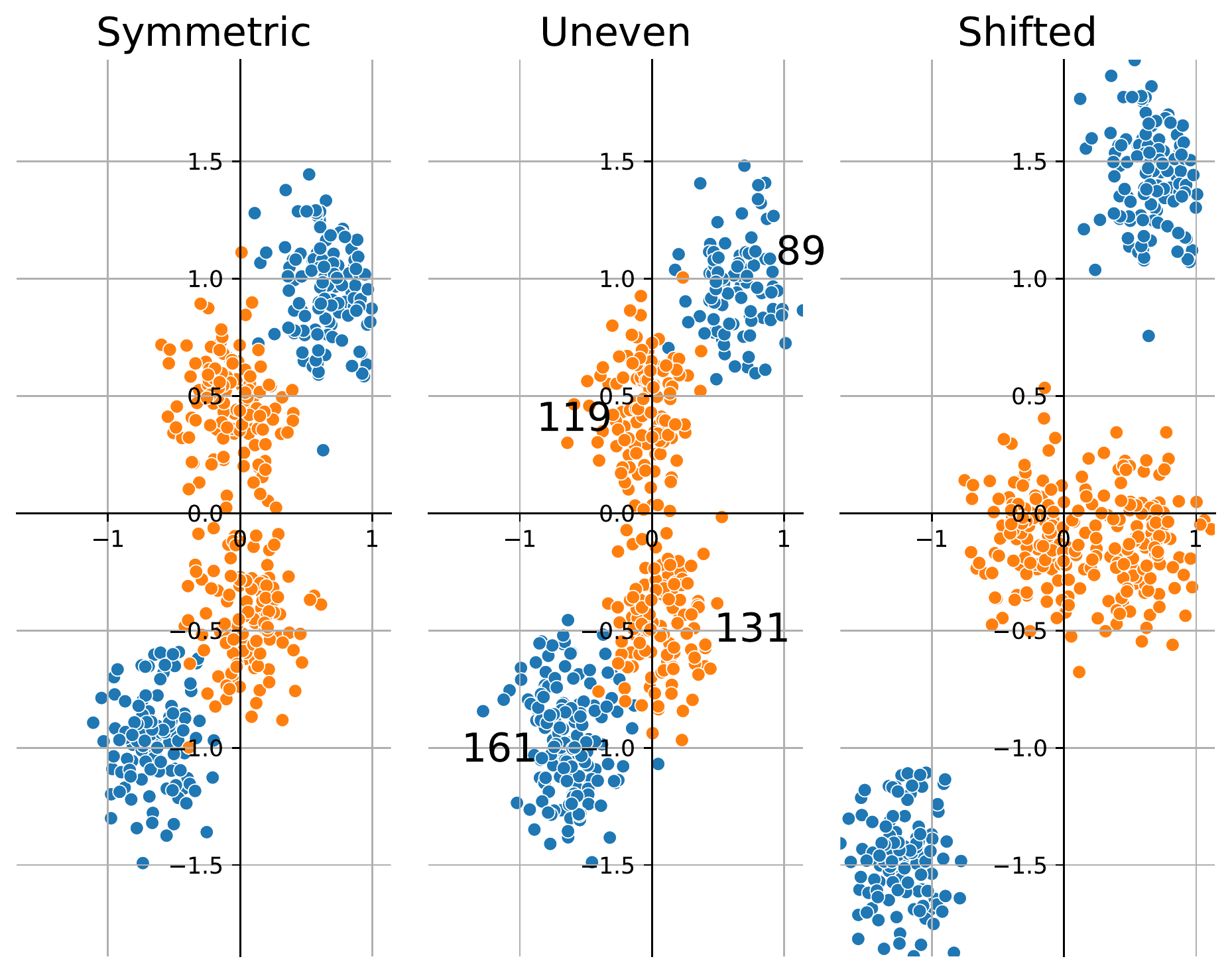}}
\vskip -0.1in
\caption{Example distributions visualized in 2 dimensions.}
\label{fig:2D_models}
\end{center}
\vskip -0.4in
\end{figure}

Figure \ref{fig:2D_models} shows all seven distributions in  2-dimensional space, and provides a visual basis for our intuitions about how the distributions compare in terms of how fully they use the space. Appendix~\ref{app:dist_viz} includes further discussion of how these distributions compare in high dimensions, leading to the following intuitions about how they should perform on reliable measures of data spread:

\begin{enumerate}
\vspace{-0.2cm}
\item The \emph{Shell}, \emph{Nested Shell}, and \emph{Sphere} distributions should all have similar scores that indicate well-spread distributions, particularly in high dimensions.
\vspace{-0.2cm}
\item The \emph{Cone} distribution should have a slightly lower score that still indicates a fairly well-spread distribution, particularly in high dimensions.
\vspace{-0.2cm}
\item All three cluster models should have similar scores that indicate poorly spread distributions, regardless of the dimensionality of the distribution.
\end{enumerate}

\section{Common Measures of Isotropy}\label{sec:common_isotropy}

Research into the effect that data spread has on NLP model performance generally relies on one or two measures of spread: average cosine similarity and I(V). We compute these values across our seven example distributions and demonstrate that both of these common measures fall short as relative measures of spread. 

\subsection{Average Cosine Similarity (ACS)}
The first and simpler of these two common measures is the average pairwise cosine similarity between word representations in an NLP model \cite{Bihani2021-pc, Ethayarajh2019-fl, Ferner2021-xe, Gao2019-zg, Liang2021-aq}. In an isotropic latent space, the expected pairwise cosine similarity between data points is zero.

\subsection{I(V)}
The second commonly used measure of spread, I(V) (sometimes called  $\gamma$), is a min/max ratio of a principal component-based partition function. I(V) was first introduced by \citet{Mu2017-ck} and has been used broadly in research involving latent space isotropy~\cite{Kaneko2020-fq,Liao2020-ll,Rajaee2021-lw, Wang2019-jn, Zhang2022-qm}. 

\citet{Mu2017-ck} build on the partition function explored by \citet{Arora2015-bu} (Equation \ref{z_partition}), which was shown to be constant in an isotropic space over all partitions, $c$,  where $v_w$ is the vector representation of word $w$. They used the full-rank set of principal components as their partitions, $C$, and proposed Equation \ref{I_v} as a measure of isotropy. I(V) ranges from zero to one, and holds the value one in a completely isotropic space.
\begin{align}
		Z_c = \sum\limits_{w}\exp(c^Tv_w) \label{z_partition}\\
		I(V)=\frac{\min_{c \in C}Z_c}{\max_{c \in C}Z_c} \label{I_v}
\end{align}
\subsection{Measure Weaknesses}\label{weak}

\begin{figure}[t]
	\begin{center}
	\centerline{\includegraphics[width = 0.88\columnwidth]{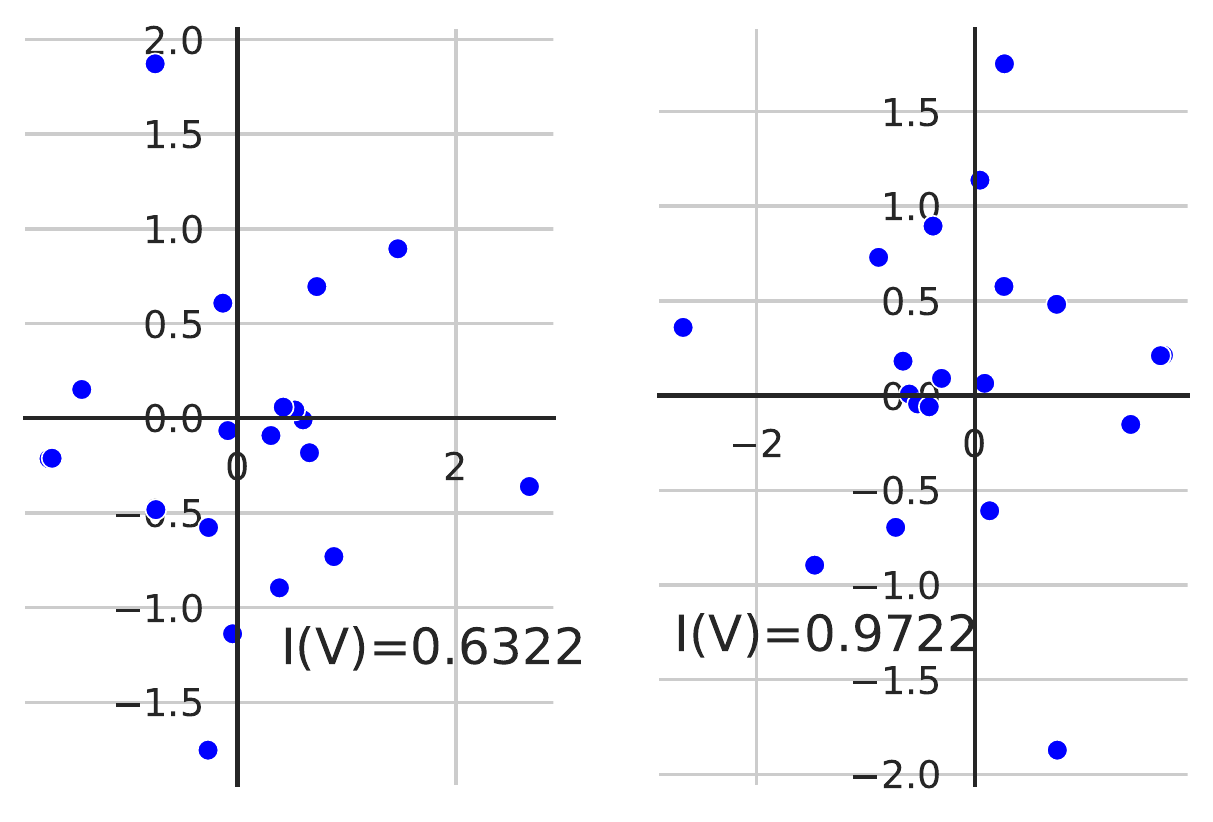}}
	\vskip -0.1in
	\caption{2D principal component projections of identical sample data with component signs switched and I(V) calculated for each set of principal components.}
	\label{fig:pc_neg}
         \end{center}
\vskip -0.3in
\end{figure}

I(V) and ACS both have characteristic results in an isotropic space ($ACS = 0$ and $I(V) = 1$), but this does not necessarily mean that they are good measures of \emph{relative} spread. In particular, we find the following weaknesses after computing their values across our example distributions (Table~\ref{common_metrics}):
\begin{itemize}
\vspace{-0.2cm}
\item{ACS is \textbf{insensitive to uneven data distributions that are symmetric across the origin}. As an example, all cluster distributions produce an ACS value similar to that of the \emph{Sphere} distribution.}
\vspace{-0.2cm}
\item{I(V) is \textbf{insensitive to the difference in spread between the \emph{Cone} and the three other spherical distributions}, particularly in high dimensions.} 
\vspace{-0.2cm}
\item{I(V) shows \textbf{inconsistent results on the cluster models} across different dimension counts \footnote{The I(V) measure is also sensitive to the random seed used to produce these distributions. The same cluster distributions with different seeds used for sampling the cluster centers and points will have I(V) scores that differ greatly.}}:
\vspace{-0.2cm}
\item{I(V) \textbf{heavily down-weights negative projections} due to the exponentiation in the partition function $Z_c$ (Equation \ref{z_partition}). In an ideally filled space, all positive and negative directions must be equally filled.}
\vspace{-0.2cm}
\item{I(V) is \textbf{sensitive to the sign of principal components}, which is arbitrary~\cite{Jolliffe2016-ui}. Figure \ref{fig:pc_neg} shows two alternative (negated) projections of randomly sampled data: when the principal components for a single distribution are negated (causing the projected data to be reflected across the origin), the resulting I(V) scores can differ greatly.}
\end{itemize}
\vskip -0.1in
\begin{table}[t]
\vskip -0.1in
	\caption{Example distribution results on \emph{Average Cosine Similarity (ACS)} and \emph{I(V)} in 2 and 100 dimensions. Full results shown in Appendix~\ref{app:results}}
        \vskip 0.1in
	\label{common_metrics}
        \begin{center}
        \begin{small}
        \begin{sc}
	\begin{tabular}{l|cc|cc}
		\toprule
		Example  & \multicolumn{2}{c|}{2D}  & \multicolumn{2}{c}{100D}           \\
		Distribution & ACS     & I(V)    & ACS     & I(V)\Bstrut\\
		\hhline{-|--|--}
		\emph{Shell} & 0.0027 & 0.9737  & 0.0007 & 0.9988\Tstrut\\
		\emph{Nested Shell} & 0.0027 & 0.9801 & 0.0007 & 0.9982 \\
            \emph{Sphere} & 0.0036 & 0.9613 & 0.0007 & 0.9988 \\
		\emph{Cone} & 0.8176 & 0.9469 & 0.5119 & 0.9944 \\
		\emph{Symm. Clust.} & 0.0020 & 0.7888 & 0.0007 & 0.9822 \\
		\emph{Shifted Clust.} & 0.0109 & 0.9295 & 0.0012 & 0.8326 \\
		\emph{Uneven Clust.} & 0.0249 & 0.7797 & 0.0043 & 0.9819 \\
            \hhline{-|--|--}
            \emph{Normal} & 0.0022 & 0.9969 & 0.0008 & 0.9988\Tstrut\\
		\bottomrule
	\end{tabular}
        \end{sc}
        \end{small}
        \end{center}
        \vskip -0.15in
\end{table}

\section{Alternative Measures of Spread} \label{sec:measures}

To better address the issues described in Section \ref{weak}, we explored a variety of new options for measuring spread in high dimensional space. Here we present two measures based on principal components/eigenvalues and four entropy-based measures.

In addition to the six measures proposed below, we explored a measure that considered the Euclidean distances between a distribution's first and $k$th nearest neighbors, and a Gaussian approximation of KL-Divergence. These measures are excluded from the main body as they did not provide new or interesting results when applied to our example distributions, but their definitions, results, and discussion can be found in Appendix~\ref{app:addtl_measures}.

\subsection{Principal Component Measures}

Principal components (eigenvectors) describe all of the orthogonal directions for a given dataset, and the associated eigenvalues describe how much of the dataset's variance can be explained along each individual direction. These eigenvalues can be used as a measure of how evenly a latent space's variance is spread along each axis. In an ideal, well-spread space, each principal component will explain an equal amount of the dataset's variance, resulting in all eigenvalues being roughly equal.  

\begin{figure}[t]
	\begin{center}
	\centerline{\includegraphics[width = \columnwidth]{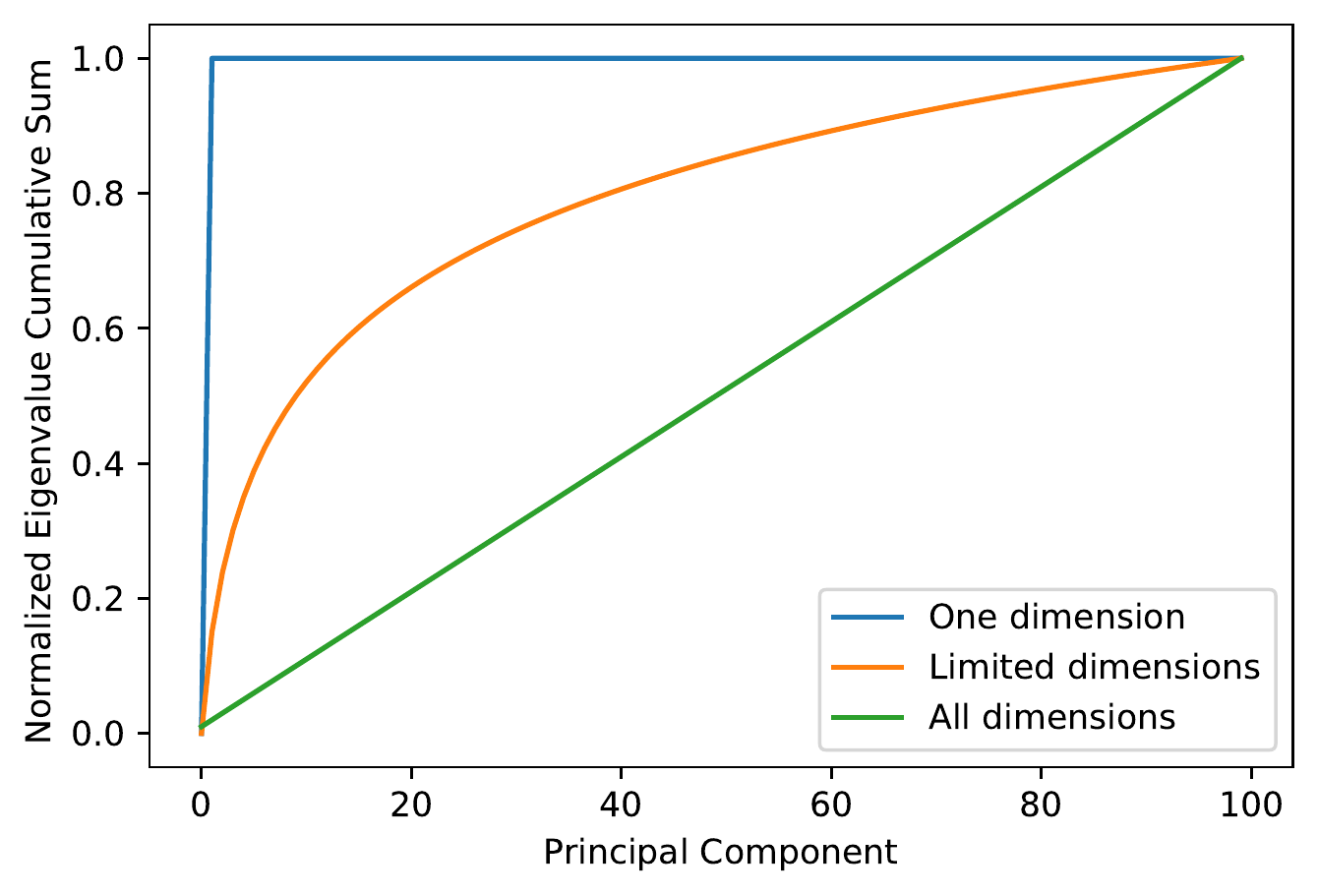}}
	\caption{Cumulative Sum of Eigenvalues for latent spaces in which variance is explained on one dimension, unevenly across many dimensions, and evenly across all dimensions.}
	\label{eigensum_demo}
        \end{center}
\vskip -0.3in
\end{figure}

\textbf{Eigenvalue Ratio (ER)} \quad
We first propose to compute the ratio of the largest and smallest eigenvalues as in Equation~\ref{eq:eigenratio}, where $C$ is the set of all $d$ principal components in a $d$-dimensional space, and $\lambda_c$ is the eigenvalue associated with component $c$. In an ideally filled latent space, this ratio  is equal to one; if one or more dimensions is poorly used, the value will approach zero. Note that this measure provides no way to differentiate between the case of one poorly utilized dimension from the case in which many (or even most) dimensions are largely unused.
\begin{align}
ER = \frac{\min_{c \in C} \lambda_{c}}{\max_{c \in C} \lambda_{c}}\label{eq:eigenratio}
\end{align}
\textbf{Eigenvalue Early Enrichment (EEE)} \quad
A related approach is to instead consider the cumulative sum of the eigenvalues (which are sorted largest to smallest by convention). Figure \ref{eigensum_demo} provides a visual example of the approach.

In Equation~\ref{eq:EEE}, we calculate the area between the cumulative sum curve, $X_{EEE}$, and the ideal linear sum, $Y_{ref}$, as a proportion of the total space available above that linear sum line (with $v = \sum_{C}\lambda_c$ as the total variance in the $d$-dimensional distribution). Well-spread distributions show an EEE value close to zero, while poorly spread distributions approach a value of one. Unlike the previous approach, EEE is able to differentiate distributions that utilize different fractions of the available latent space.
\begin{align}
EEE &= \frac{AUC(X_{EEE}-Y_{ref})}{\frac{1}{2}dv} \label{eq:EEE}
\end{align}

\subsection{Entropy Ratio Measures} \label{entropy_measures}

In information theory, the concept of \emph{entropy} quantifies the amount of uncertainty involved in predicting an outcome, which can be mapped to the concept of \emph{spread}: with poorly spread data, certainty is high and entropy is low; with well-spread data, uncertainty is high, as is entropy. 
Shannon entropy \cite{shannon1951prediction} is defined for continuous data as in Equation \ref{shannon} and has been proposed, in the field of astronomy, as a measure of the isotropy of the universe \cite{Pandey2016-yc}. 
\begin{equation}\label{shannon}
H(p) = \int p(x) \log p(x)dx
\end{equation}
For a fixed variance, this parametric definition is maximized by the normal distribution \cite{Arizono1989-nz, Beirlant1997-jo}, which matches our intuition (from Section \ref{spread_dist}) that a multivariate normal is a good candidate for a distribution that ideally fills the available space. Thus, the two empirical Shannon entropy approximations discussed below will be compared to a multivariate normal with equal variance to create relative measures of spread.

\textbf{Vasicek Ratio Mean Squared Error (VRM)} \quad
Our first entropy-based measure of data spread builds on the Vasicek entropy approximation~\cite{vasicek1976test}, which is limited to univariate data and rests on the notion that ordered points will be evenly spaced in a highly entropic space. The approximation is presented in Equation \ref{eq:vasicek}, and considers pairs of points that are separated by a fixed interval, $m$, where $m$ is a positive integer smaller than $n/2$, $n$ is the total number of data points, $x_j = x_1$ if $j<1$, and $x_j = x_n$ if $j>n$. 
\begin{align}
H_{vas} &= \frac{1}{n}\sum_{i=1}^{n}\log \frac{n}{2m}(x_{i+m}-x_{i-m}) \label{eq:vasicek} 
\end{align}
To create a relative measure of data spread, we consider a ratio of the empirical value for a given distribution to the theoretical value for our reference normal distribution, $\ln(\sqrt{2\pi e}\sigma^{2})$\cite{Arizono1989-nz}. This ratio will equal one for normally distributed data points and will approach zero for poorly spread data \footnote{This ratio is similar to Pielou's evenness index from the biodiversity literature~\cite{Pielou1966-gc}.}. We compute the mean squared error (MSE) from the target ratio of one to create a multi-dimensional measure (Equation~\ref{eq:vac_multi}). 
\begin{align}
VRM&= \frac{1}{d}\sum\limits_{i=1}^{d}(1-\frac{H_{vas}}{\ln(\sqrt{2\pi e}\sigma^{2})})^2 \label{eq:vac_multi}
\end{align}

\textbf{Nearest Neighbor Entropy Ratio (NNR)} \quad
A multivariate continuous entropy approximation rests on the notion that highly entropic spaces will maximize the minimum distance between points and their nearest neighbors \cite{Beirlant1997-jo,Leonenko2008-yw,Sablayrolles2018-zh}. Equation \ref{eq:nnr} presents this nearest neighbor approximation as defined by~\citet{Beirlant1997-jo}, in which  $\rho_{i}$ is the distance of element $i$ to its nearest neighbor, and $e$ is the Euler constant. 
\begin{align}
H_{NN}&=\sum_{i=1}^{n}\ln(\rho_{i})+\ln(2n)+e \label{eq:nnr}
\end{align}
To create a relative measure, we simulate a $d$-dimensional multivariate normal distribution and empirically calculate $H_{NN}$ for this distribution as our theoretical maximum. We use this in a ratio as in the previous section to create our proposed measure of spread\footnote{This measure is similar to the Clark-Evans measure of dispersion from spatial statistics \cite{clark1979generalization}.}. A fully used space will produce a ratio close to one and a poorly used space will produce a value approaching zero.
\begin{equation}
NNR = \frac{H_{NN}(X)}{H_{NN}(Y)}, \quad Y\sim N(0,\sigma^2)
\end{equation}

\subsection{KL-Divergence Measures}
KL-Divergence (Equation \ref{eq:KL-div}) provides an established framework for comparing the entropy of two distributions. We compare our example distributions directly to a multivariate normal distribution to create a relative measure of spread. 
Below, we describe two measures of spread based on empirical KL-Divergence approximations.
\begin{equation}\label{eq:KL-div}
KL = \int_{x}p(x)\log\frac{p(x)}{q(x)}
\end{equation}

\textbf{Discrete KL-Divergence Mean Squared Error (DKLM)} \quad
KL-Divergence can be computed for finite data sets by discretizing the data as in Equation~\ref{eq:KL-discrete}. We estimate univariate $p(X')$ and $q(Y')$ by binning observed and reference data ($X$, $Y$) into $k$ bins along each dimension (with $k=30$) and compute the MSE from a reference of $KL_{disc} = 0$ over all $d$ dimensions, as in Equation~\ref{eq:KL-SSE}.
\begin{align}
KL_{disc}=\sum_{i=1}^{k}p(x_{i}')\log\frac{p(x_i')}{q(x_i')} \label{eq:KL-discrete}\\
DKLM = \frac{1}{d}\sum\limits_{i=1}^{d}(KL_{disc_i})^2 \label{eq:KL-SSE}
\end{align}

\textbf{Nearest Neighbor KL-Divergence (NNKL)} \quad
Finally, we propose a KL-Divergence approximation based on comparing the distributions of nearest neighbor distances in our observed and reference distributions. We present Equation~\ref{eq:NNKL}, a modified version of the approximation derived in \citet{Perez-Cruz2008-tn}, which relies only on nearest neighbor distances to point $x_i$ in our observed and reference distributions, $s(x_i)$ and $r(x_i)$ respectively, the number of dimensions, $d$, and the number of data points, $n$ (observed) and $m$ (reference). 
\begin{align}
NNKL=\frac{d}{n}\sum_{i=1}^{n}\max(0,\log\frac{r_{k}(x_{i})}{s_{k}(x_{i})}) \label{eq:NNKL}
\end{align}

\section{Results}
Here, we evaluate each of the alternative measures proposed in Section~\ref{sec:measures} as applied to the seven example distributions described in Section~\ref{sec:sample_models} and demonstrate that all measures better reflect our expectations as relative measures of spread. We review the results for each of our proposed measures below\footnote{Patterns described in the main body hold for 10 and 50 dimensions as well, and complete results can be found in Appendix \ref{app:results}.}, and finally apply our two strongest measures to a pre-trained Word2Vec model.

\subsection{Principal Component Measures}

\begin{table}[t]
	\caption{Example distribution results on \emph{Eigenvalue Ratio (ER)} and \emph{Eigenvalue Early Enrichment (EEE)} in 2 and 100 dimensions.}
	\label{tab:pc_measure_results}
        \vskip 0.15in
        \begin{center}
        \begin{small}
        \begin{sc}
	\begin{tabular}{l|cc|cc}
		\toprule
		Example & \multicolumn{2}{c|}{2D} & \multicolumn{2}{c}{100D}           \\
		Distribution &
		ER     & EEE      & ER     & EEE\Bstrut\\
		\hhline{-|--|--}
		\emph{Shell} & 0.9021 & 0.0129 & 0.7845 & 0.0361\Tstrut\\
		\emph{Nested Shell} & 0.8552 & 0.0195 &  0.7542 & 0.0422 \\
		\emph{Sphere} & 0.9118 & 0.0115 & 0.7889 & 0.0353\\
		\emph{Cone} & 0.4173 & 0.1028 & 0.0090 & 0.0458 \\
		\emph{Symm. Clust.} & 0.1124 & 0.1995 & 0.0001 & 0.5394 \\
		\emph{Shifted Clust.} & 0.0915 & 0.2081 &  0.0007 & 0.6746 \\
		\emph{Uneven Clust.} & 0.1102 & 0.2004 &  0.0001 & 0.5397 \\
            \hhline{-|--|--}
            \emph{Normal} & 0.7910 & 0.0292 &  0.7804 & 0.0362\Tstrut\\
		\bottomrule
	\end{tabular}
\end{sc}
\end{small}
\end{center}
\vskip -0.1in
\end{table}

\begin{figure}[t]
\vskip 0.2in
\begin{center}
	\centerline{\includegraphics[width = \columnwidth]{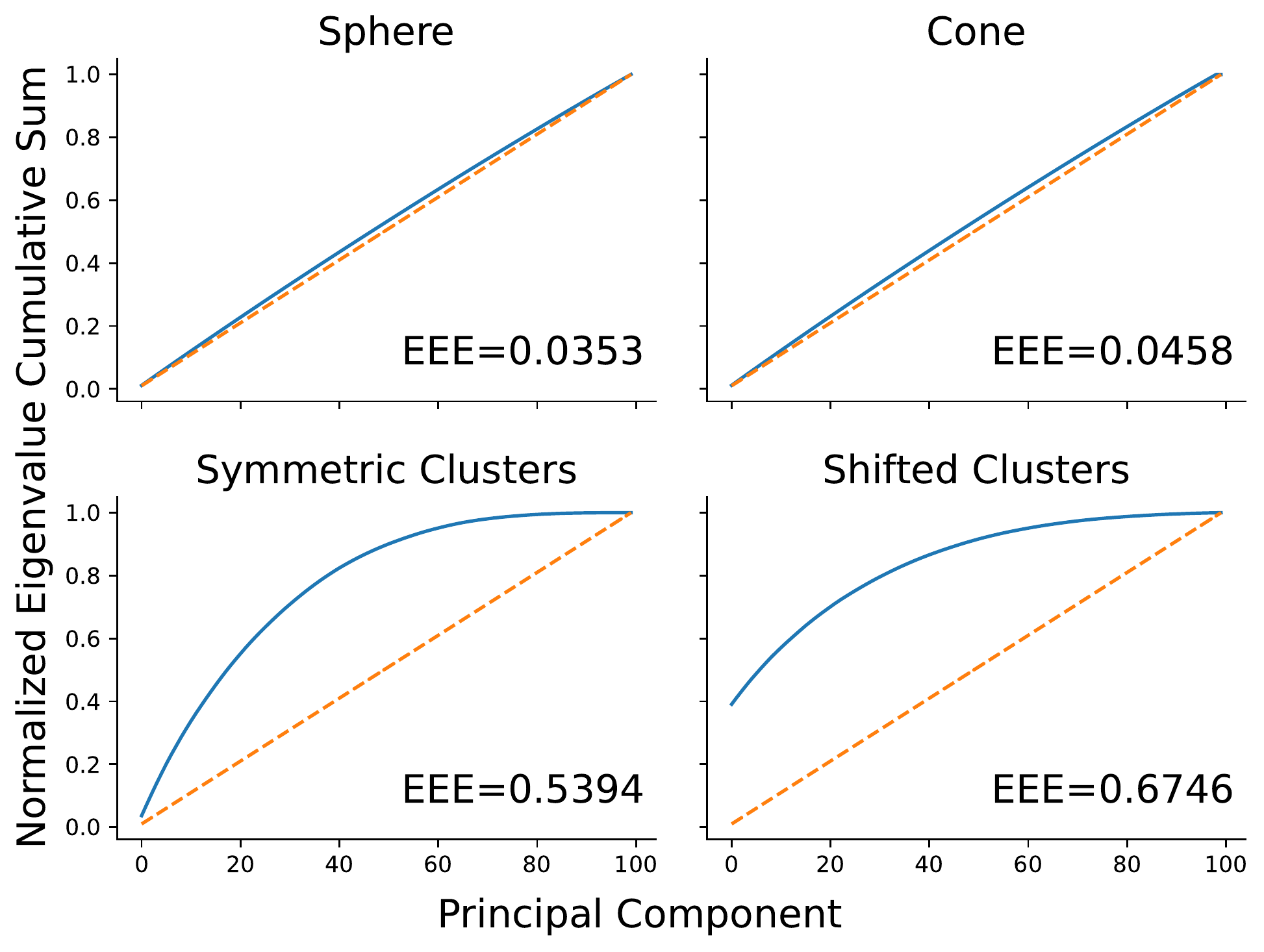}}
	\caption{\emph{Eigenvalue Early Enrichment (EEE)} scores for \emph{Sphere}, \emph{Cone}, \emph{Symmetric Clusters}, and \emph{Shifted Clusters} distributions}
	\label{fig:EEE_AUC}
\end{center}
\vskip -0.3in
\end{figure}

Our proposed ER measure considers the ratio between the largest and smallest eigenvalues of a data distribution (i.e. the explained variance for the first and last principal component), where a ratio of one indicates a well-spread distribution (Equation~\ref{eq:eigenratio}). The EEE score instead considers the area under the curve of the cumulative sum of eigenvalues as a proportion of the total area available, where a score of zero indicates a well-spread distribution (Equation~\ref{eq:EEE}). 

As seen in Table \ref{tab:pc_measure_results}, both measures show a gradient of spread across our example distributions in 2 and 100 dimensions, with the \emph{Sphere} and \emph{Shell} distributions most evenly spread and the three cluster distributions least well-spread. 

Although both of these measures seem to perform well on our example distributions, the EEE score better agrees with our expectations for a relative measure of spread, since it considers more than just the first and last eigenvalue. Indeed, Table~\ref{tab:pc_measure_results} shows a disagreement between the ER score and the EEE score on the 100-dimensional cluster distributions which can be explained by examining the eigenvalue cumulative sum curves in Figure~\ref{fig:EEE_AUC}. The ER score penalizes the \emph{Symmetric Clusters} distribution for having a very small last eigenvalue, while the EEE score more accurately reflects the more gradual decrease over all eigenvalues when compared to the \emph{Shifted Clusters} distribution.

\subsection{Univariate Entropy Measures}\label{subsec:entropy}

VRM considers a ratio of the Vasicek entropy approximation between an observed distribution and a normal distribution (Equation~\ref{eq:vasicek}) and DKLM considers the KL-divergence between a discretized observed distribution and a normal distribution (Equation~\ref{eq:KL-discrete}). We calculated the MSE over all dimensions to translate these into multivariate measures, and thus zero indicates a well-spread distribution for both measures.

\begin{figure}[t]
\begin{center}
	\centerline{\includegraphics[width = \columnwidth]{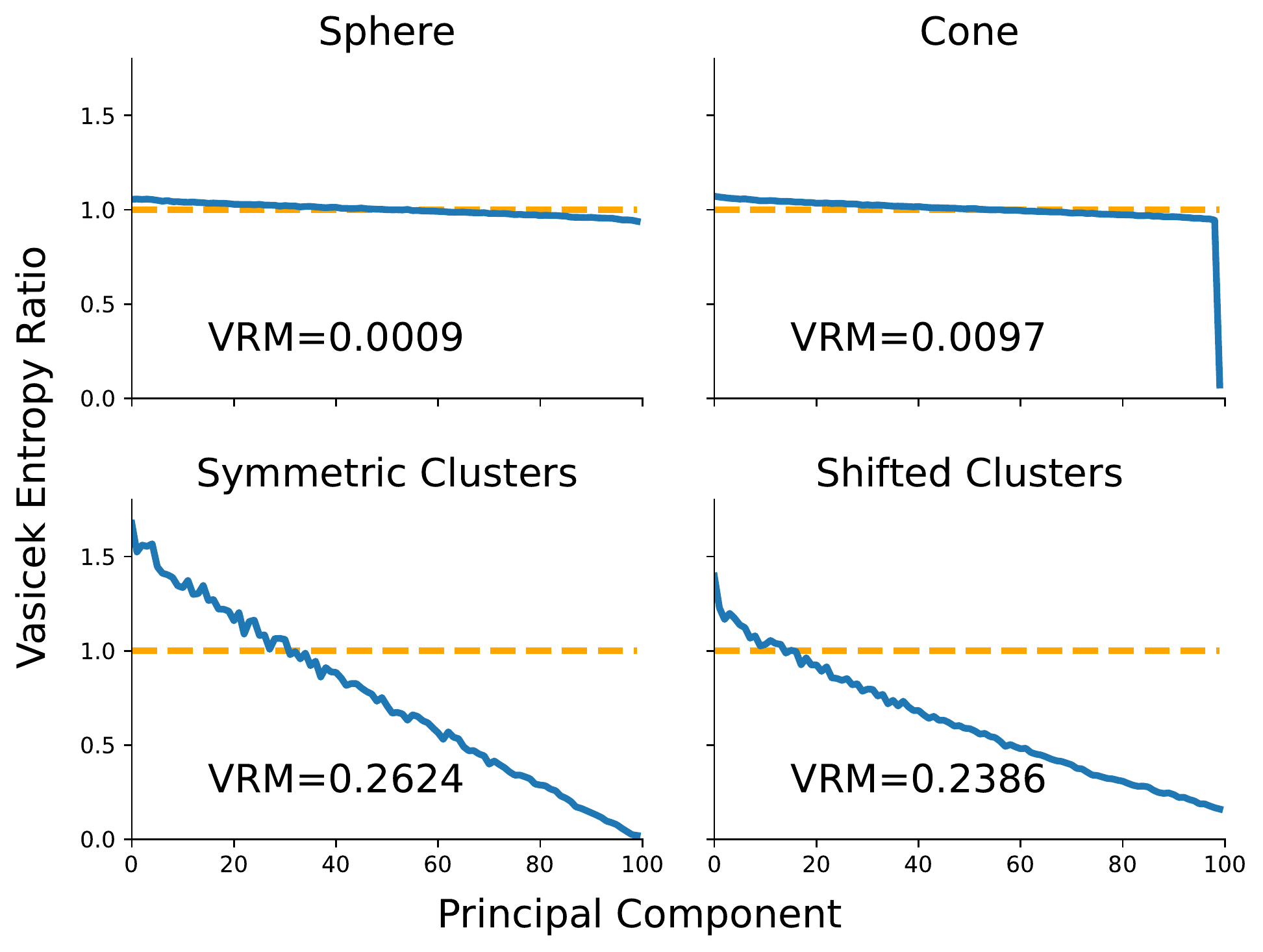}}
 \vskip -0.1in
	\caption{Vasicek Entropy Ratio for 100-dimensional distributions in blue, with the reference line in orange. In a well-spread distribution, the ratio would always be one, and the VRM score would be zero.}
	\label{fig:vas_SSE}
 \end{center}
 \vskip -0.3in
\end{figure}

\begin{figure}[t]
\begin{center}
	\centerline{\includegraphics[width = \columnwidth]{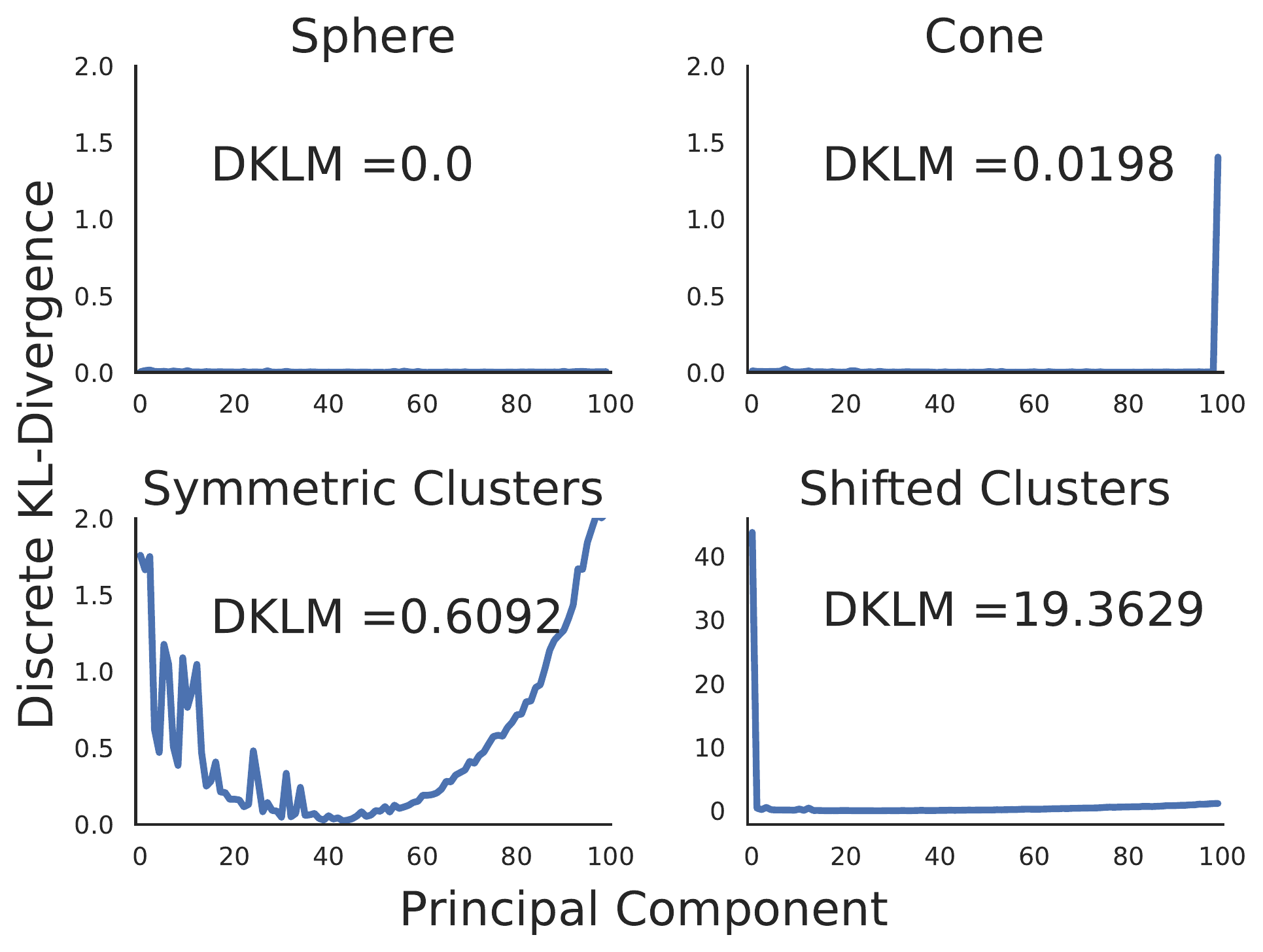}}
 \vskip -0.1in
	\caption{Discrete KL-Divergence for 100-dimensional models. In a well-spread distribution the KL-Divergence and DKLM scores would both be zero.}
	\label{fig:disc_KL_SSE}
 \end{center}
 \vskip -0.3in
\end{figure}

Both VRM and DKLM behave as we would expect for quality relative measures of spread, particularly in high dimensions (see Table \ref{tab:univ_ent_results} and Figures~\ref{fig:vas_SSE} and \ref{fig:disc_KL_SSE} \footnote{To ensure that the measures are not influenced by covariance between dimensions, we projected our distributions onto their principal components before computing our univariate entropy measures. This is reflected in the negative slopes and u-shaped distributions in Figures~\ref{fig:vas_SSE} and \ref{fig:disc_KL_SSE}, respectively.}). Both measures produce small MSE for the \emph{Sphere}, \emph{Shell}, and \emph{Nested Shell} distributions, a larger MSE for the \emph{Cone} distribution, and an even larger MSE for the cluster distributions.

\begin{table}[t]
\vskip -0.1in
\caption{Example distribution results for \emph{Vasicek Ratio MSE (VRM)} and \emph{Discrete KL-Divergence MSE (DKLM)} in 2 and 100 dimensions.}
\label{tab:univ_ent_results}
\vskip 0.15in
\begin{center}
\begin{small}
\begin{sc}
	\begin{tabular}{l|rr|rr}
		\toprule
		  Example & \multicolumn{2}{c|}{2D} & \multicolumn{2}{c}{100D}           \\
            Distribution &
		\multicolumn{1}{c}{VRM}     & \multicolumn{1}{c|}{DKLM}     & \multicolumn{1}{c}{VRM}     & \multicolumn{1}{c}{DKLM}\Bstrut\\
		\hhline{-|--|--}
		\emph{Shell} & 0.1891 & 0.1349 & 0.0010 & 0.0000\Tstrut\\
		\emph{Nested Shell} & 0.0465 & 0.0774 & 0.0014 & 0.0019 \\
		\emph{Sphere} & 0.0106 & 0.0169  & 0.0009 & 0.0000\\
		\emph{Cone} & 0.0686 & 0.0266 & 0.0097 & 0.0295 \\
		\emph{Symm. Clust.} & 0.1545 & 0.1434 & 0.2624 & 0.6200 \\
		\emph{Shifted Clust.} & 0.2143 & 0.5392 & 0.2386 & 20.5810 \\
		\emph{Uneven Clust.} & 0.1637 & 0.1470 &  0.2574 & 0.6333 \\
            \hhline{-|--|--}
            \emph{Normal} & 0.0047 & 0.0190 &  0.0010 & 0.0000\Tstrut\\
		\bottomrule
	\end{tabular}
\end{sc}
\end{small}
\end{center}
\vskip -0.2in
\end{table}

The \emph{Shifted Clusters} chart in Figure \ref{fig:disc_KL_SSE} shows how just a few components dominate the DKLM score for this distribution, which is possible because KL-Divergence does not have an upper limit. Indeed, our KL-divergence measures (DKLM and NNKL) are functions of the number of data points and the number of dimensions, such that the range of scores for distributions in high dimensions is much larger than in lower dimensional distributions\footnote{Table \ref{nn_ent_results} provides the strongest example of this issue, with the NNKL scores for the 100-dimensional cluster distributions being almost 300x the scores of their 2-dimensional counterparts.}. An ideal measure of spread should allow meaningful comparison between distributions with different dimensionality.

\subsection{Nearest Neighbor Entropy Measures}
NNR considers a ratio between the nearest-neighbor entropy approximation of the observed data and that of a multivariate normal distribution (Equation~\ref{eq:nnr}), and NNKL considers a nearest-neighbor approximation of the KL-divergence between the observed data and a multivariate normal distribution. In a well-spread distribution, we would expect an NNR of one, and an NNKL of zero.

Table \ref{nn_ent_results} shows that, although these measures are once again effective at identifying the less-complete spread of the cluster distributions, the \emph{Cone} distribution receives similar scores to the \emph{Sphere} and \emph{Shell} distributions, while the scores for the \emph{Nested Shell} distribution indicate a less complete spread, especially in high dimensions. 

\begin{figure}[t]
\begin{center}
	\centerline{\includegraphics[width = \columnwidth]{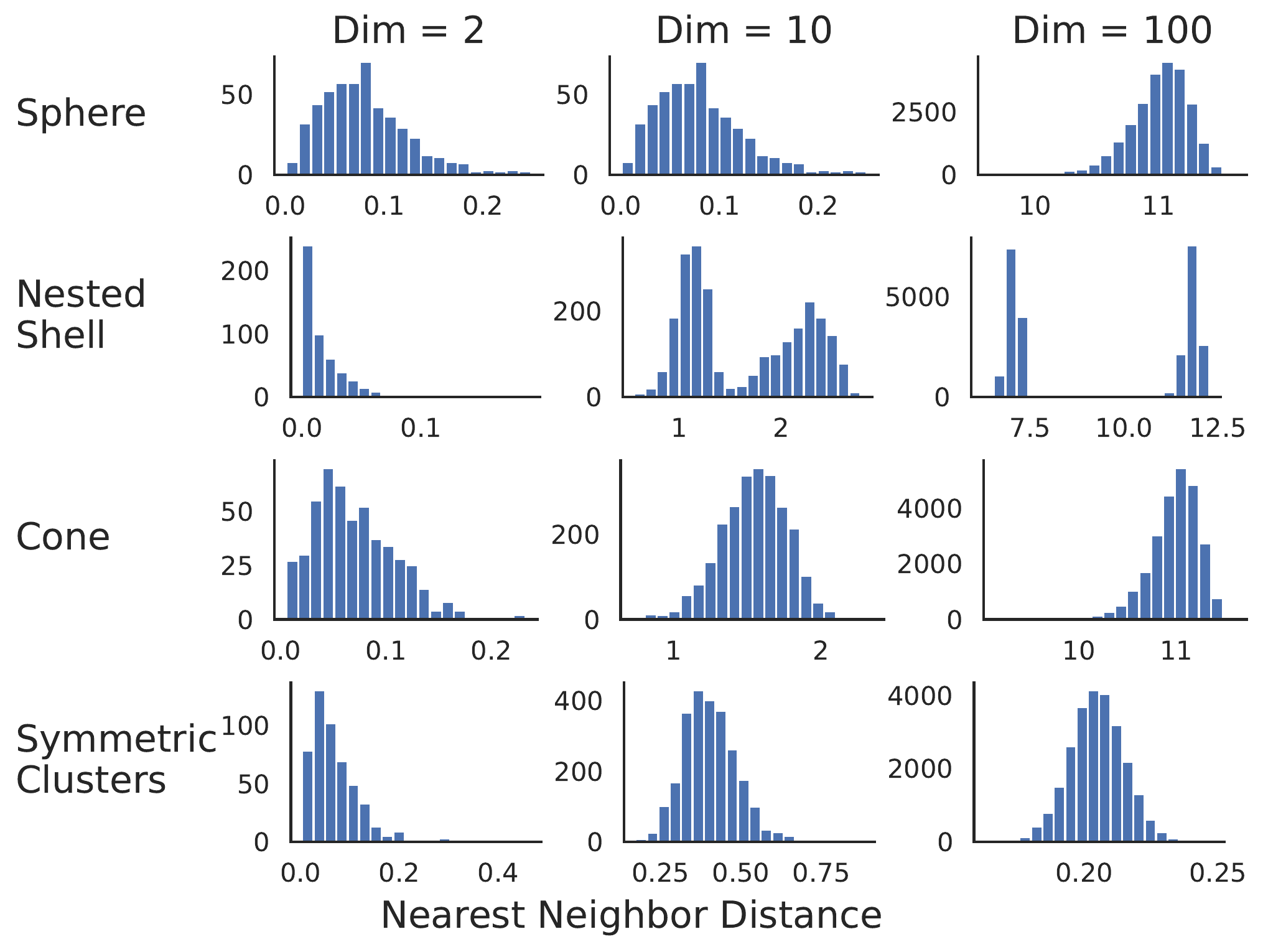}}
 \vskip -0.1in
	\caption{Histograms of nearest neighbor distances for our \emph{Sphere}, \emph{Nested Shell}, \emph{Cone}, and \emph{Symmetric Clusters} example distributions in 2, 10, and 100 dimensions.}
	\label{model_nns}
\end{center}
\vskip -0.3in
\end{figure}

Figure \ref{model_nns} shows histograms of the nearest neighbor distances in our example distributions, and demonstrates that these distances are very similar for the \emph{Sphere} and \emph{Cone} distributions as the number of dimensions increases, while the nearest neighbor distances for the \emph{Nested Shell} distribution become increasingly separated for the two ``rings'' in the distribution. This suggests that measures that rely wholly on Euclidean distances may not be as robust as some of the other measures we have examined.

\begin{table}[t]
\vskip -0.1in
\caption{Example distribution results on \emph{Nearest Neighbor Entropy Ratio (NNR)} and \emph{Nearest Neighbor KL-Divergence (NNKL)} in 2 and 100 dimensions.}
\label{nn_ent_results}
\vskip 0.15in
\begin{center}
\begin{small}
\begin{sc}
	\begin{tabular}{l|rr|r@{\hspace{1.3\tabcolsep}}r}
		\toprule
		  Example & \multicolumn{2}{c|}{2D} & \multicolumn{2}{c}{100D}           \\
		Distribution & \multicolumn{1}{c}{NNR} & \multicolumn{1}{c|}{NNKL} & \multicolumn{1}{c}{NNR} & \multicolumn{1}{c}{NNKL}\Bstrut\\
		\hhline{-|--|--}
		\emph{Shell} & 0.1323 & 7.4573 & 1.0024 & 0.2532\Tstrut\\
		\emph{Nested Shell} & 0.3996 & 5.2696  & 0.9887 & 13.2125 \\
		\emph{Sphere} & 0.9749 & 0.9267  & 1.0023 & 0.2558 \\
		\emph{Cone} & 0.9350 & 1.2659 & 1.0022 & 0.2853 \\
		\emph{Symm. Clust.} & 0.8971 & 1.4874  & 0.7135 & 394.5467 \\
		\emph{Shifted Clust.} & 0.8312 & 1.9226  & 0.6882 & 429.2692 \\
		\emph{Uneven Clust.} & 0.9015 & 1.4215  & 0.7134 & 394.7038 \\
            \hhline{-|--|--}
            \emph{Normal} & 0.9973 & 0.7260  & 1.0001 & 1.1751\Tstrut\\
		\bottomrule
	\end{tabular}
\end{sc}
\end{small}
\end{center}
\vskip -0.1in
\end{table}

\subsection{Word2Vec Embedding Performance}

The work in this paper is largely motivated by creating better metrics for comparing the spread in natural language processing models' latent spaces. Thus, as a final exploration, we compare the performance of a pre-trained Word2Vec model (300-dimensions trained on the Google News corpus, \citealp{mikolov_2013, mikolov2013distributed}) on our two strongest candidate measures, EEE and VRM, along with the current common measures, ACS and I(V). We follow the methods used in generating our example distributions and randomly sample 75000 ($250d$) word embeddings from this pre-trained model to use in calculating our measures of data spread. To examine whether these measures can capture a gradual increase in data spread when applied to a real (not simulated) latent space, we add random uniform noise to the pre-trained embeddings and calculate the EEE, VRM, ACS, and I(V) scores at each addition. 

Table~\ref{tab:w2v_scores} shows that the original pre-trained Word2Vec model falls somewhere between our example \emph{Cone} distribution and our \emph{Symmetric Clusters} distribution according to all four of these measures of spread. Additionally, we see that all four measures successfully reflect the gradual increase in data spread as random noise is added to the pre-trained word embeddings. This does not discount the issues with ACS and I(V) that we raised in Section~\ref{sec:common_isotropy}, but further supports the use of EEE and VRM as relative measures of data spread that do not suffer from the same weaknesses.

\begin{table}[t]
\vskip -0.1in
\caption{Pre-trained Word2Vec Results on Early Eigenvalue Enrichment (EEE), Vasicek Ratio MSE (VRM), Average Cosine Similarity (ACS), and I(V) measures with gradual addition of noise sampled from uniform distributions with the listed range.}
\label{tab:w2v_scores}
\vskip 0.15in
\begin{center}
\begin{small}
\begin{sc}
	\begin{tabular}{l|rrrr}
		\toprule 
		\multicolumn{1}{c|}{Noise} &
		   \multicolumn{1}{c}{EEE} & \multicolumn{1}{c}{VRM} & \multicolumn{1}{c}{ACS}  & \multicolumn{1}{c}{I(V)}\Bstrut\\
		\hhline{-|----}
		\emph{None} & 0.4058 & 0.1525 & 0.1317 & 0.9251\Tstrut\\
		$\pm$0.001 & 0.4058 & 0.1525 & 0.1319 & 0.9251 \\
		$\pm$0.01 & 0.4049 & 0.1514 & 0.1280 & 0.9257\\
		$\pm$0.05 & 0.3861 & 0.1320 & 0.1147 & 0.9359\\
		  $\pm$0.1 & 0.3375 & 0.0961 & 0.0933 & 0.9534\\
		$\pm$0.3 & 0.1492 & 0.0216 & 0.0346 & 0.9883\\
		$\pm$0.5 & 0.0800 & 0.0066 & 0.0192 & 0.9952\\
            $\pm$1 & 0.0441 & 0.0016 & 0.0062 & 0.9991\\
            $\pm$3 & 0.0367 & 0.0010 & 0.0022 & 0.9996\\
		\bottomrule
	\end{tabular}
\end{sc}
\end{small}
\end{center}
\vskip -0.2in
\end{table}

\section{Conclusion}

In this work, we have examined methods for quantifying how completely data fills a latent space. We demonstrated that the metrics commonly being used to quantify this usage are insufficient as relative measures of data spread, and we proposed six alternative measures of data spread. Of our proposed measures, all improved upon the commonly used measures when applied to seven synthetic data distributions, and we present one principal component measure and one entropy-based measure, EEE (Early Eigenvalue Enrichment) and VRM (Vasicek Ratio MSE) respectively, as our strongest proposed measures.

Future work that builds on our findings includes the reassessment of previous methods and the development of new methods for increasing data spread in NLP models using these two proposed measures. We expect that the application of reliable measures of data spread in this manner will contribute to the general understanding of NLP and other neural network models, by further defining the geometric properties associated with improved benchmarking performance.

\bibliography{biblio}
\bibliographystyle{icml2023}

\newpage

\appendix

\section{Distribution Definitions}\label{app:dists}
\paragraph{Shell}\hfill\newline
A relatively simple way to sample points from the surface of a $d$-dimensional unit hypersphere is to sample from a multivariate standard normal and normalize the length of the associated vectors \cite{Muller1959-wb}. Here, $X^{shell}_i$ is the vector for one of the $250d$ points sampled from our $d$-dimensional shell. 
\begin{align}
		X^{shell}_{i} &= \frac{X_i}{\parallel X_i \parallel}, \quad X_i \sim N(0,I_d) \label{eq:shell-sample}
\end{align}
\paragraph{Nested Shell}\hfill\newline
To create a nested shell, that is, a shell within a shell, we adjust the radius of half of our shell points to be $\frac{1}{2}$ instead of $1$.
\begin{align}
r^{nest}_{i}&=
\left\{ \begin{array}{ll}
1 &i\leq 250d/2 \\
1/2 & i>250d/2
\end{array} \right.\\
X^{nest}_{i} &= r^{nest}_{i} X^{shell}_{i} 
\end{align}
\paragraph{Sphere}\hfill\newline
We can sample from a uniformly filled unit hypersphere by taking the points from our \emph{Shell} distribution and randomizing their distance from the origin between 0 and the radius of the hypersphere, $r$. However, due to the exponential relationship between the radius and the volume of a $d$-dimensional hypersphere (as seen in Equations \ref{eq:SA} and \ref{eq:vol},  $V_d = f(r^d)$)\footnote{This exponential relationship is directly related to the discussion in Section~\ref{sec:spread_geom}, since points that are distributed uniformly within a high-dimensional hypersphere will end up largely in the neighborhood of the hypersphere's radius.}, we cannot sample the distance of each point directly from a $U(0,r)$ distribution without causing points to be more densely concentrated around the origin. Thus, we invert this exponent when sampling our distance from the origin, as seen in Equation \ref{eq:sphere-radius}. 
\begin{align}
		l^{sphere}_{i} &= l^\frac{1}{d}_i, \quad l_i \in L \sim U(0,r) \label{eq:sphere-radius}\\
		X^{sphere}_{i} &= l^{sphere}_{i} X^{shell}_{i} \label{eq:sphere-sample}
\end{align}
\paragraph{Cone}\hfill\newline
A 3-dimensional cone can be described as a continuous series of circles (2-dimensional spheres), stacked along a third dimension (the cone's height), where the radius of each circle is a function of the distance from the origin, $l$, and the angle/width of the cone, $\theta$: as we move further from the origin (as $l$ grows), the radius of each circle increases according to the width of the cone.

Analogously, a $d$-dimensional hypercone can be imagined as a series of $(d-1)$-dimensional hyperspheres that are continuously stacked along the $d$th dimension. Again, the radius of the $i$th stacked sphere, $r^{stack}_i$, is a function of $l^{cone}_i$ (the sphere's distance from the origin) and $\theta$ (the angle/width of the cone)\footnote{We chose to define $\theta=1/\sqrt{d}$ as it produced distributions that more consistently demonstrated the strengths and weaknesses of common and proposed measures of spread across different numbers of dimensions.}. 
As in the case of the \emph{Sphere} distribution, we invert the exponential relationship between volume and distance from the origin to uniformly sample within the cone as in Equation~\ref{eq:rcone}.
\begin{align}
		l^{cone}_{i} &= l^\frac{1}{d}_i,  \quad l_i \in L\sim U(0,r) \label{eq:rcone}\\
		r^{stack}_{i} &= l^{cone}_{i}\tan (\theta)
\end{align}\\
We then sample from an $(d-1)$-dimensional sphere of radius $r^{stack}_i$ as in Equations \ref{eq:sphere-radius} and \ref{eq:sphere-sample}. The points sampled from these $(d-1)$-dimensional spheres ($X^{stack}_i$) are concatenated with the distance from the origin ($l^{cone}_i$) to produce $d$-dimensional vectors as in Equation \ref{eq:hypercone_sample}.
\begin{align}\label{eq:hypercone_sample}
X^{cone}_i &= (l^{cone}_i, X^{stack}_i)
\end{align}
\paragraph{Clusters}\hfill\newline
Cluster distributions were created by randomly sampling $d$ cluster centers, mirroring these centers over the origin to create a total of $2d$ cluster centers, and randomly sampling around each center $\mu_j$ to create clusters. In Equation \ref{eq:symm_clust} we use the floor function to ensure integer division so that the \emph{Symmetric Clusters} distribution sampled 125 points around each cluster center. 
\begin{align}
		\mu^{symm}_j &=  
		\left\{ \begin{array}{ll}
        \;\;\;\mu_j &j\leq d \\
        -\mu_{j/2} & j>d
        \end{array} \right. , \quad \mu_j \sim U(-1,1)\\
		X^{symm}_{i} &\sim N(\mu^{symm}_{\lfloor{}i/125\rfloor}, \min (1/d, 0.2))\label{eq:symm_clust}
\end{align}\\
To break symmetry, we amend Equation \ref{eq:symm_clust} in two ways. For the \emph{Shifted Clusters} distribution, we randomly shifted each cluster center and again sampled 125 points around each cluster center.
\begin{align}
		\mu^{shift}_i &= \mu^{symm}_i+S_i, \quad S_i \sim U(0,1)
\end{align}\\
For the \emph{Uneven Clusters} distribution, we randomized the number of points in each symmetric cluster, such that the clusters in each mirrored cluster pair, ($\mu^{symm}_{i}$, $\mu^{symm}_{{i+d}}$), have $k$ and $(250-k)$ points, respectively, and $k$ is sampled from a $U(0,250)$ distribution.

\section{Example Distributions in High Dimensions}\label{app:dist_viz}

\subsection{Geometry}\label{app:geom}
While not central to our work, there are several non-intuitive geometric characteristics that will come up when discussing data spread in high dimensional spaces. Here we provide a brief description of two particular characteristics that are often included in defining the \emph{curse of dimensionality}.

First, the volume and surface area of a hypersphere (with a fixed radius) approach zero as the number of dimensions grows above $\sim$7. Second, the volume approaches zero more quickly than the surface area, which results in almost all of the points within a uniformly filled hypersphere being concentrated on a very thin shell of that hypersphere. The equations for surface area ($S_d(r)$) and volume ($V_d(r)$) in $d$ dimensions are shown in Equations \ref{eq:SA} and \ref{eq:vol} respectively.
\begin{align}
S_d(r) = \frac{2\pi^{d/2}}{\Gamma(d/2)}r^{d-1}\label{eq:SA} \\
V_{d}(r) = S_{d}\frac{r}{d}\label{eq:vol}
\end{align}
In Equation \ref{eq:SA} the gamma function in the denominator will dominate as $d$ grows, bringing the surface area to zero as the number of dimensions increases. The additional $d$ factor in the denominator of the volume function causes $V_d(r)$ to approach zero even more quickly than $S_d(r)$ as $d$ grows, with the result that data points are increasingly concentrated in the neighborhood of a hypersphere shell with a fixed radius as the number of dimensions grows \footnote{It should be noted that this effect is only observed when considering a bounded space, such as a hypersphere, as can be seen by comparing the histograms of the vector norms of a hypersphere and a normal distribution in Figure~\ref{fig:model_norms}.}

\subsection{Visualizing the Distributions}\label{sec:spread_geom}
\begin{figure}
	\centering
	\begin{minipage}{0.225\linewidth}
		\centering
		\includegraphics[width = \textwidth]{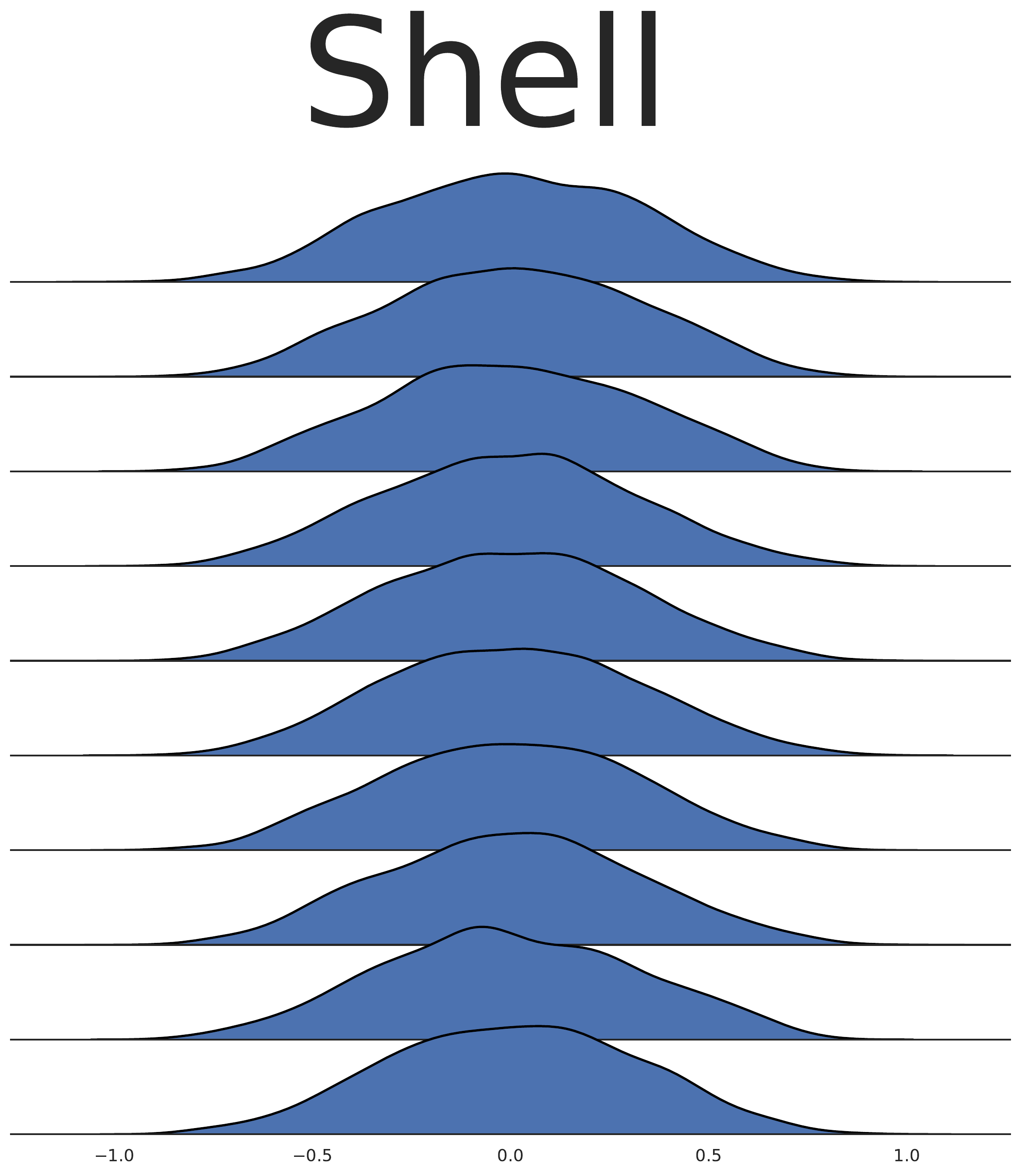}
	\end{minipage}
	\begin{minipage}{0.225\linewidth}
		\centering
		\includegraphics[width = \textwidth]{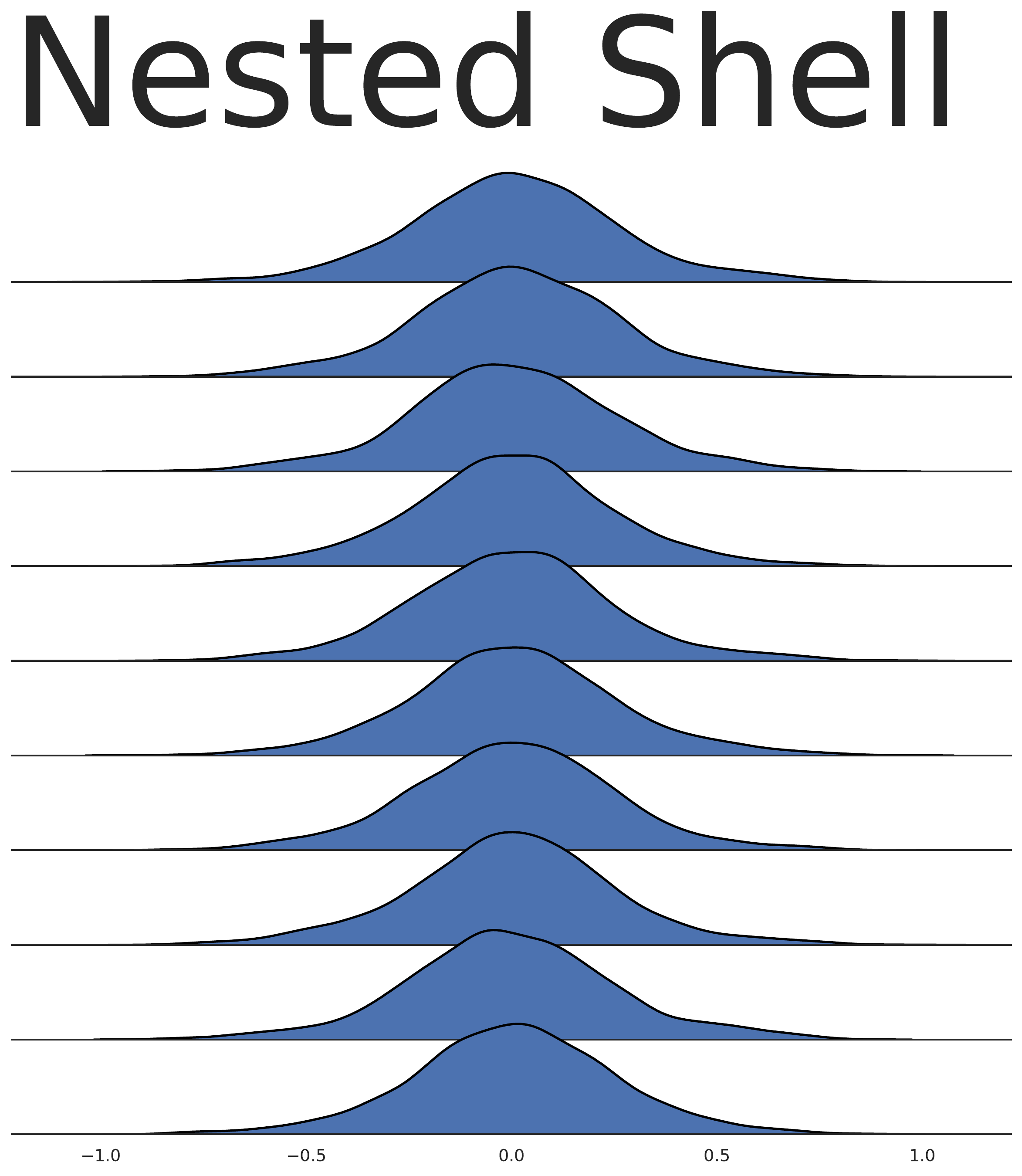}
	\end{minipage}
	\begin{minipage}{0.225\linewidth}
		\centering
		\includegraphics[width = \textwidth]{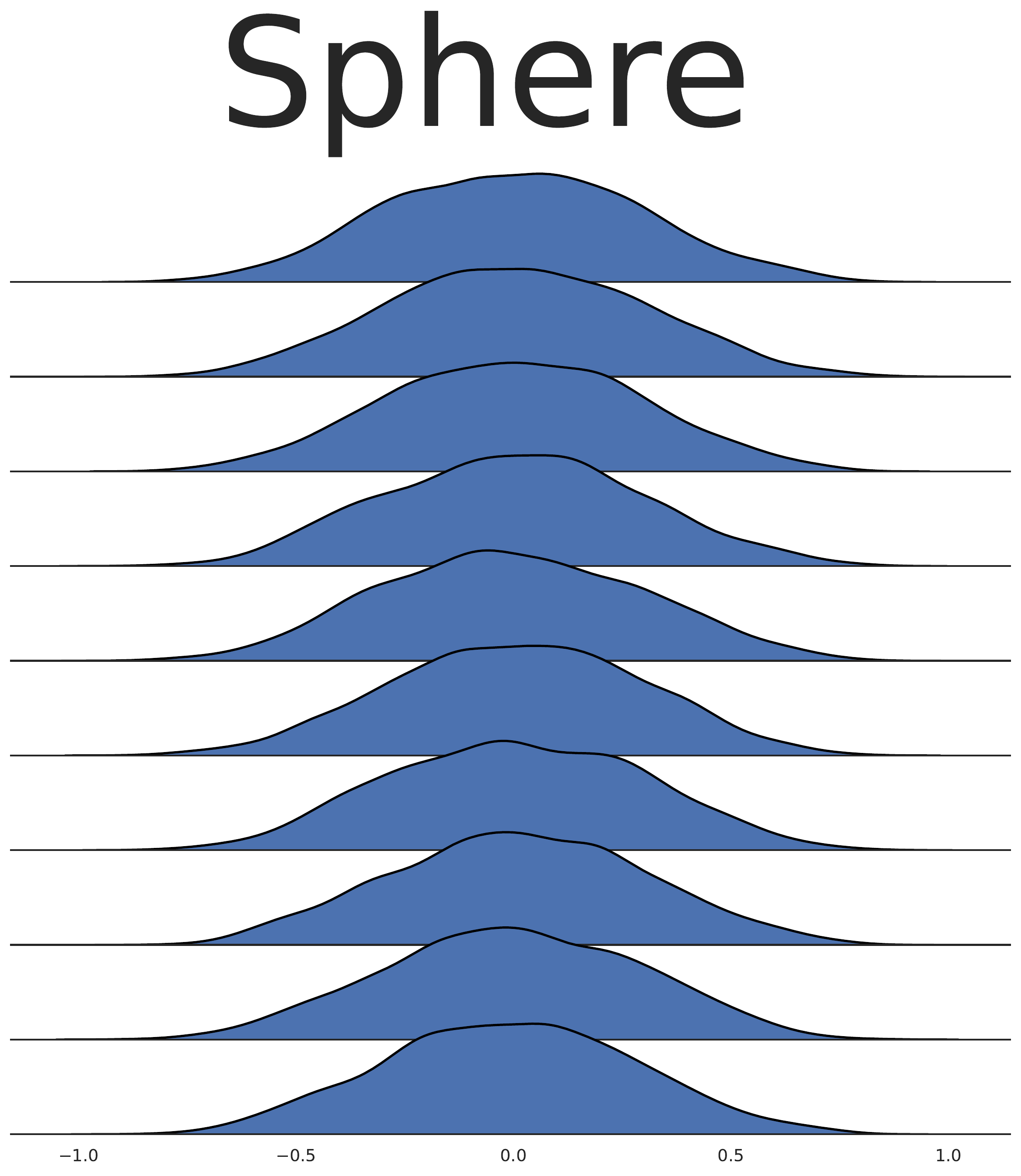}
	\end{minipage}
	\begin{minipage}{0.225\linewidth}
		\centering
		\includegraphics[width = \textwidth]{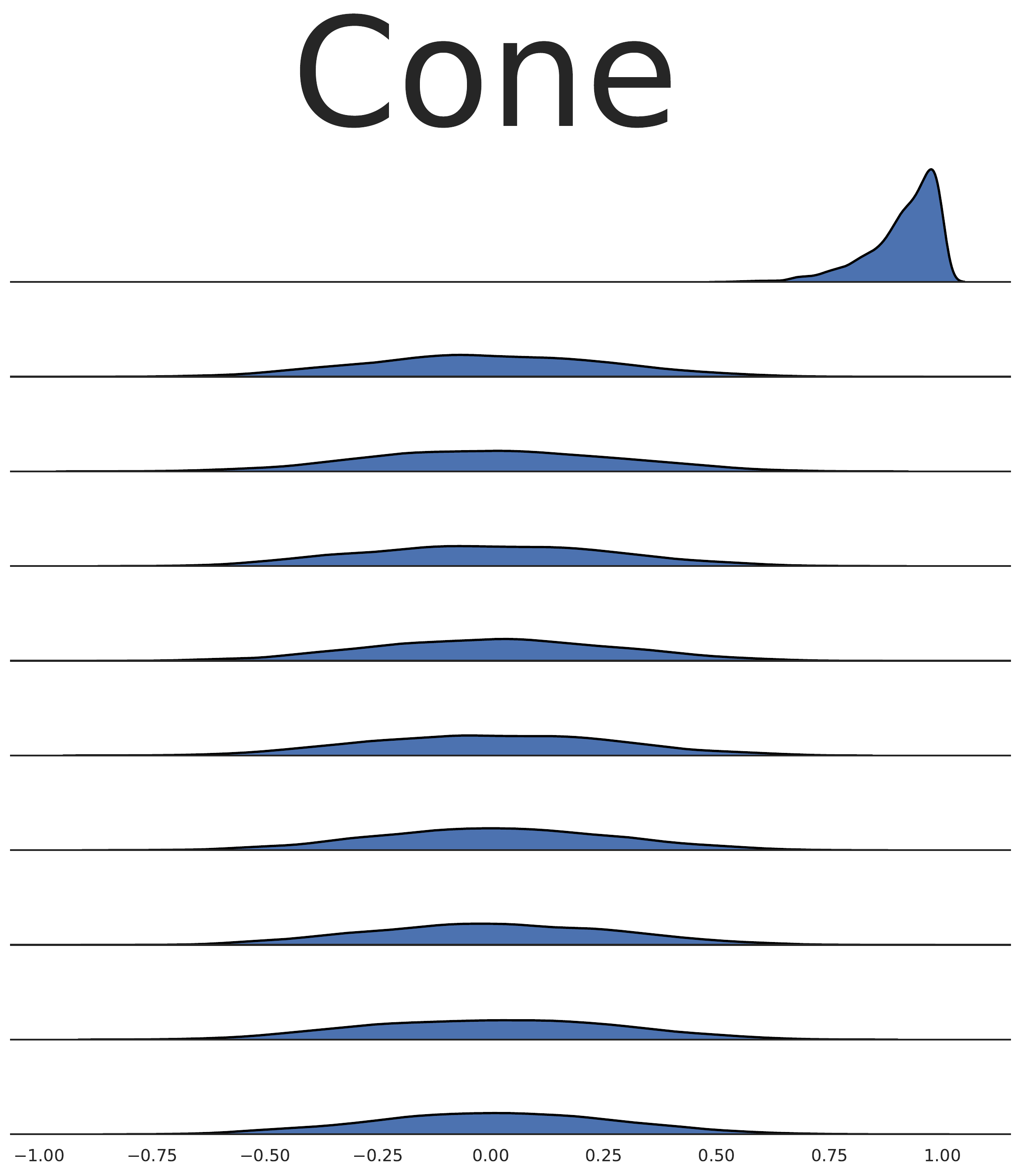}
	\end{minipage}\par\vspace{1\baselineskip}
	\begin{minipage}{0.225\linewidth}
		\centering
		\includegraphics[width = \textwidth]{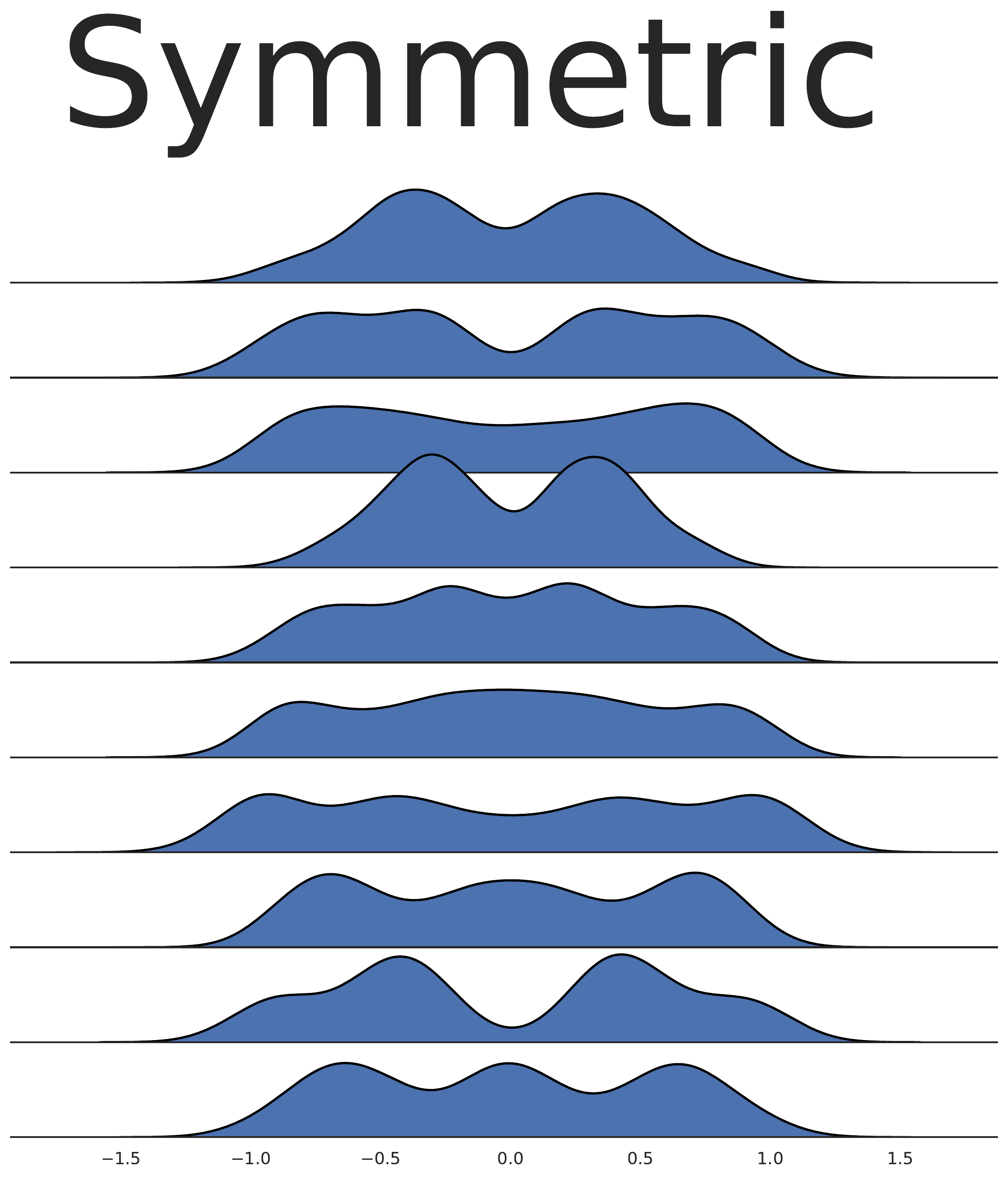}
	\end{minipage}
	\begin{minipage}{0.225\linewidth}
		\centering
		\includegraphics[width = \textwidth]{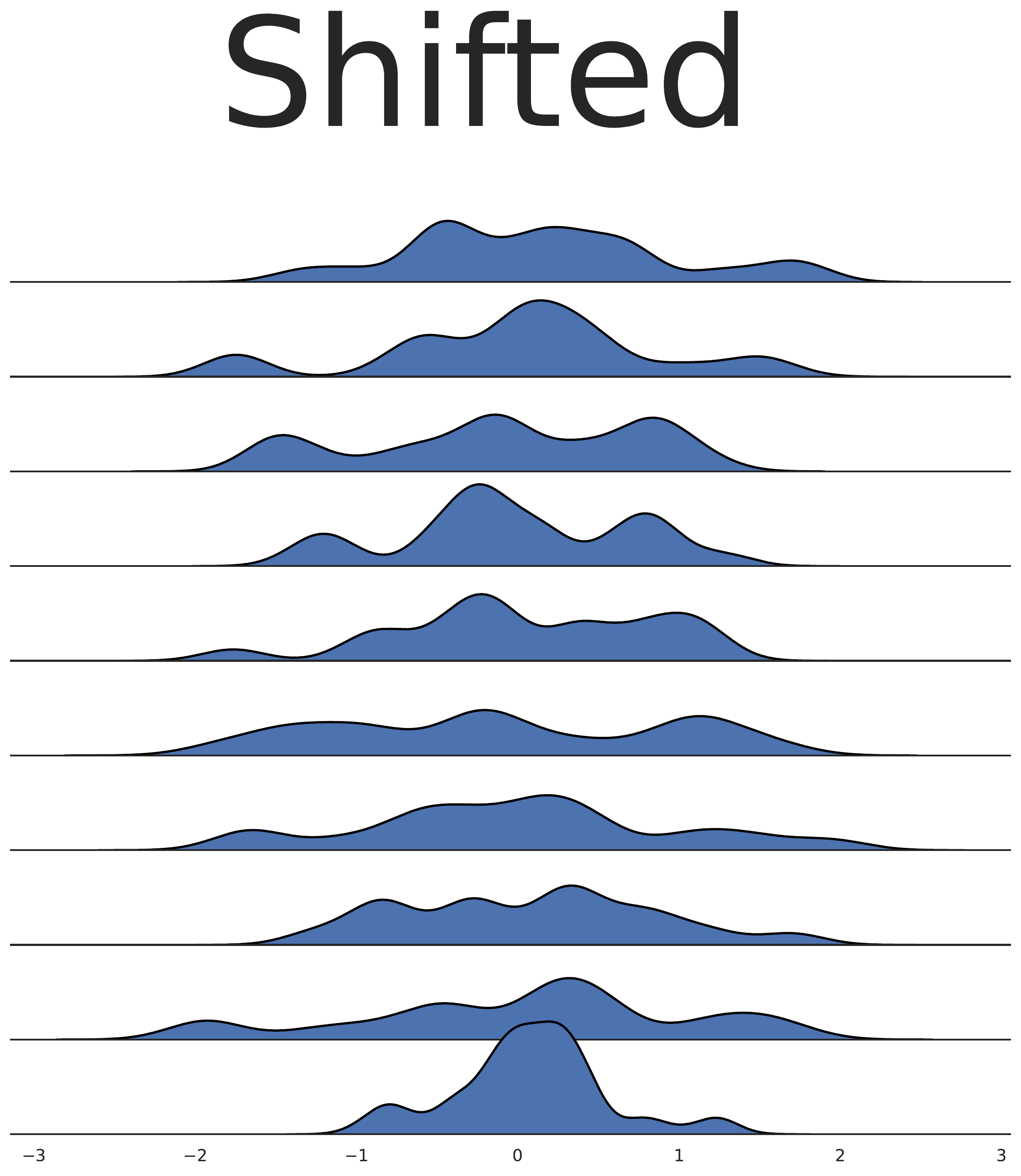}
	\end{minipage}
	\begin{minipage}{0.225\linewidth}
		\centering
		\includegraphics[width = \textwidth]{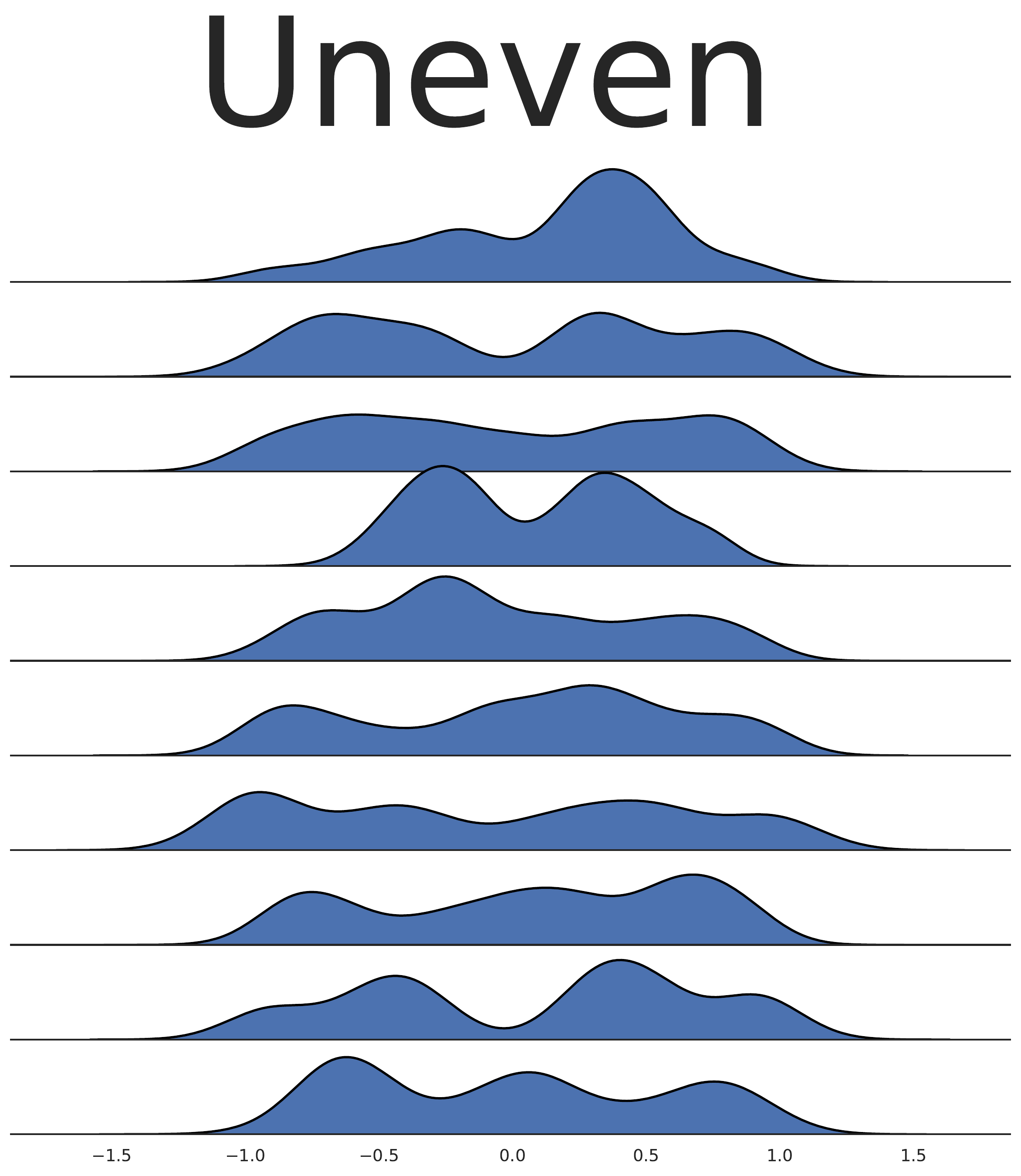}
	\end{minipage}
\caption{Density plots for 10-Dimensional example distributions show distribution of data values along each dimension; Top row shows the four spherical models; Bottom row shows the three cluster models.}	
\label{fig:ridgelines}
\end{figure}

Figure \ref{fig:ridgelines} shows density plots of the raw data values along each dimension in the 10-dimensional distributions. Even in just ten dimensions, the distribution of data among the three spherical distributions (\emph{Shell}, \emph{Nested Shell}, \emph{Sphere}) are quite similar, and even the \emph{Cone} distribution is fairly similar with the exception of the $d$th dimension along which the cone points. As discussed in Section~\ref{app:geom}, this is a consequence of the data points being more densely concentrated on the shell of a fixed-radius hypersphere as the number of dimensions grows.

The density plots for the cluster distributions help to demonstrate the effect of the transformations between the \emph{Symmetric Clusters} distribution and the \emph{Shifted} and \emph{Uneven Clusters} distributions. They also make clear the stark difference between the data uniformity of the spherical distributions and the cluster based distributions.

\begin{figure}
	\centering
	\includegraphics[width = \columnwidth]{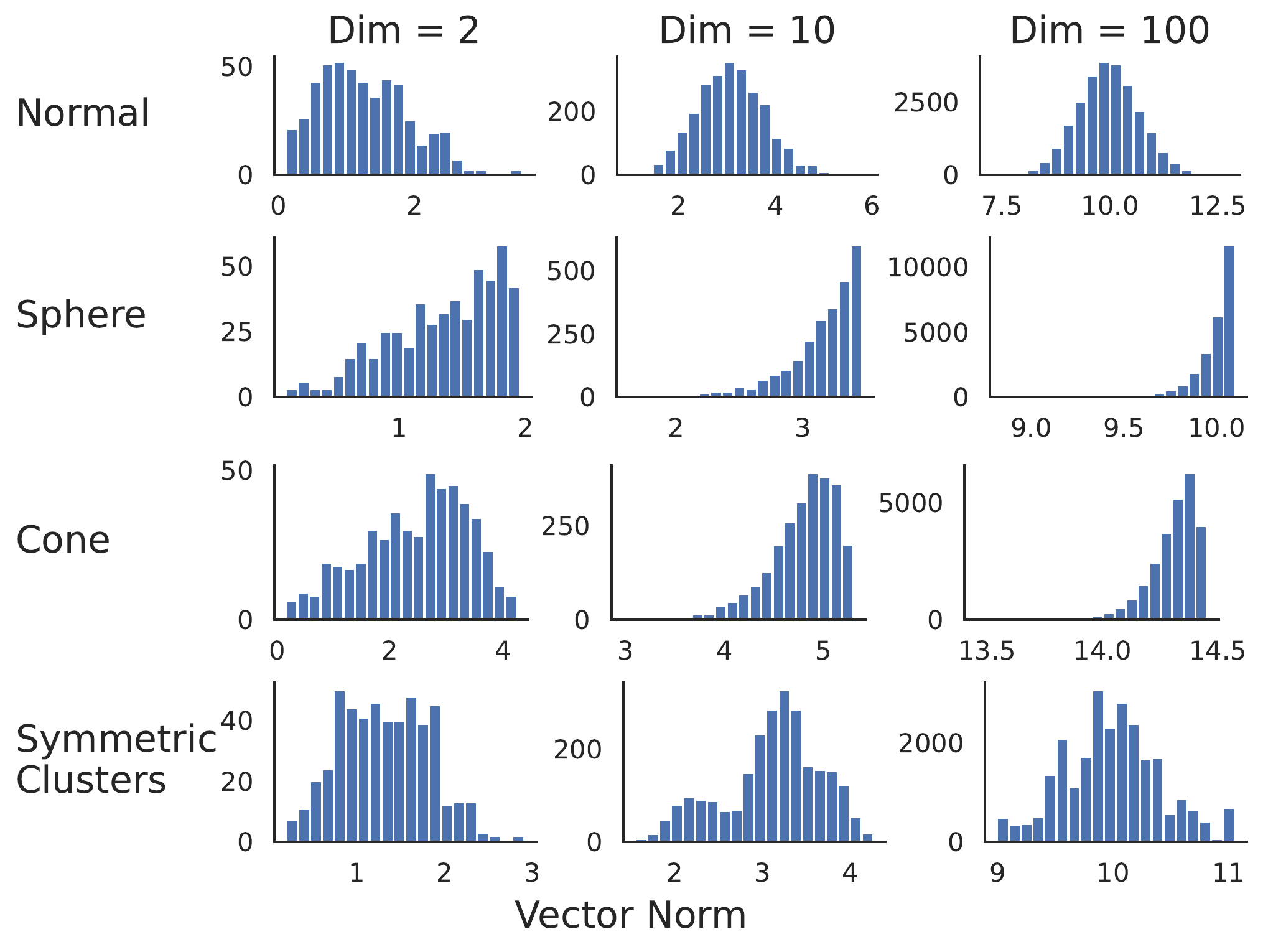}
	\caption{Histograms of vector norms for our reference \emph{normal} distribution and our \emph{Sphere}, \emph{Cone}, and \emph{Symmetric Clusters} example distributions in 2, 10, and 100 dimensions}
	\label{fig:model_norms}
\end{figure}

Figure \ref{fig:model_norms} shows histograms of vector norms for our \emph{Sphere}, \emph{Cone}, and \emph{Symmetric Clusters} distributions in 2, 10, and 100 dimensions, with a reference normal distribution for comparison\footnote{The \emph{Cone} distribution's vector norms are longer than the other distributions across all dimension counts. This is an artifact of holding the variance equal across all distributions, which we do to provide a more accurate comparison to the normal distribution as described in Section~\ref{entropy_measures}}. The first item of note is that, as the number of dimensions grows, 
the shape of the \emph{Sphere} and \emph{Cone} histograms becomes increasingly similar to the expectation for the \emph{Shell} histogram (in which all vectors have the same norm by definition). Again, this follows from the characteristic described in Section \ref{app:geom}, in that points within a hypersphere are forced onto a thin shell of that hypersphere in high dimensions. 

Our cluster distributions are not affected by this characteristic of high-dimensional geometry, since each cluster is defined by a mean and variance, rather than a strict radius. This is apparent in the histograms for the \emph{Symmetric Clusters} distribution, which is largely unchanged in shape as the dimensions grow.

\section{Additional Proposed Measures of Spread}\label{app:addtl_measures}

\subsection{KNN Overlap} \label{nn_measures}
One (computationally expensive) way to simulate a fully used latent space is to maximize the smallest pairwise Euclidean distance between points in a distribution. Indeed, there are several works that build on this concept by creating loss functions designed to maximize the distance of each point to its nearest neighbors during training \cite{Liu2018-qb, Sablayrolles2018-zh}. We sought to develop a relative measure based on this concept.

In visualizing the effect of their nearest-neighbor-based loss function, \citet{Sablayrolles2018-zh} include a chart showing histograms for the first and 100th nearest neighbor distances for a sample of their data. The motivation behind this approach is that, for an evenly filled space, the distribution of Euclidean distances of the first nearest neighbors for all points will not overlap with (will be smaller than) the distribution of the distances to the 100th nearest neighbors. Alternatively, in an unevenly used space, these distributions will overlap. 

\begin{figure}[b!]
\vskip 0.2in
	\begin{center}
	\centerline{\includegraphics[width = \columnwidth]{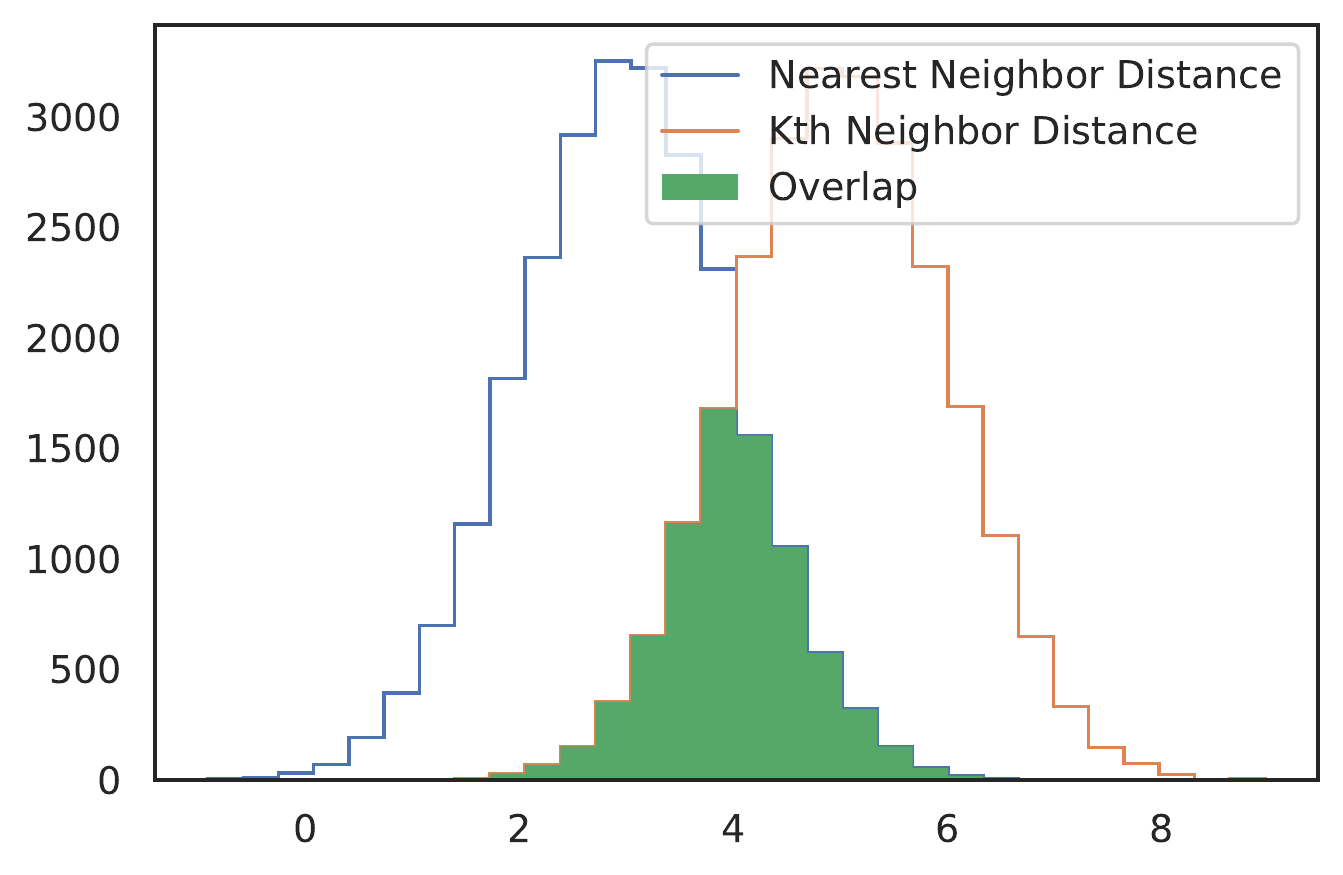}}
	\caption{Based on findings from \citet{Sablayrolles2018-zh}, histograms of Euclidean distance to first and $k$th nearest neighbors should have less overlap in a well-spread distribution than in a poorly spread distribution.}
	\label{fig:knn_ovl_demo}
        \end{center}
\vskip -0.2in
\end{figure}

Although \citet{Sablayrolles2018-zh} did not quantify this overlap, we developed a measure based on the proportion of sampled data points falling in the intersection of the distributions for the first and $k$th nearest neighbor distances, where $k=\min(5d,100)$ (for $d$ dimensions). This measure is visualized in Figure~\ref{fig:knn_ovl_demo}. Values range from zero (the expected value when for a well-spread distribution) to one (for a poorly spread distribution).

This was the only measure that we explored that didn't clearly improve on commonly used measures. As seen in Table \ref{tab:knn_ovl_results}, this measure is very small for almost all example distributions. And although the 2-dimensional distribution values reflect a reasonable relative measure of spread, these relative differences become smaller as the number of dimensions increases. We did try adjusting the value for $k$ to account for this difference, but found that the scores for the cluster distributions were particularly sensitive to the choice of $k$. 

\begin{figure}[t]
	\vskip 0.2in
        \begin{center}
	\centerline{\includegraphics[width = \columnwidth]{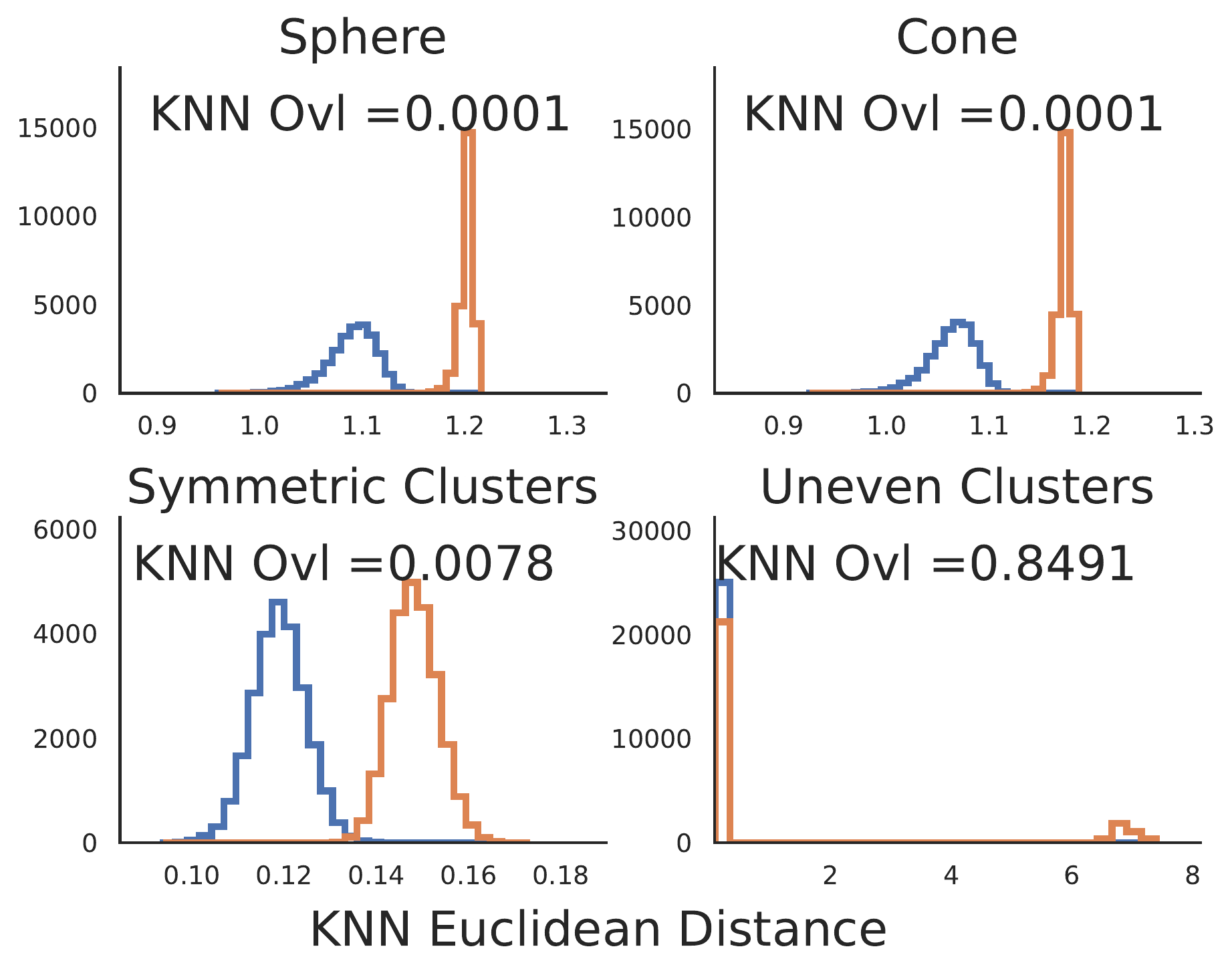}}
	\caption{KNN Overlap Proportion for 100D \emph{Sphere}, \emph{Cone}, \emph{Symmetric Clusters}, and \emph{Shifted Clusters} distributions, with the nearest neighbor distances in blue, and the 100th nearest neighbor distances in orange}
	\label{fig:knn_ovl}
        \end{center}
        \vskip -0.2in
\end{figure}

Figure \ref{fig:knn_ovl} shows that the \emph{Uneven Clusters} distribution has very high overlap with $k=100$. In the \emph{Uneven Clusters} distribution, the randomization of cluster size will cause many of the clusters to have more than the standard 125 data points used in the \emph{Symmetric Clusters} distribution. However, the randomly sized clusters still share the same tight distribution around a point, causing the 100th nearest neighbor to very frequently be quite close. Alternatively, increasing $k$ causes the \emph{KNN Overlap} score for the \emph{Symmetric Clusters} and \emph{Shifted Clusters} distributions to drop to zero, since their clusters are are all exactly 125 points (meaning that a point's 126th nearest neighbor is almost guaranteed to be much farther away than its first nearest neighbor).

\begin{table}
\caption{Example distribution results for KNN Overlap.}
\vskip 0.15in
\begin{center}
\begin{small}
\begin{sc}
    \begin{tabular}{l|cccc}
    	\toprule
    	Example Dist. &
    	2D  & 10D & 50D   & 100D\Bstrut\\
    	\hhline{-|----}
    	\emph{Shell} & 0.0120 & 0.0000 & 0.0000 & 0.0000\Tstrut\\
    	\emph{Nested Shell} & 0.0480 & 0.0924 & 0.0000  & 0.0000  \\
    	\emph{Sphere} & 0.0240 & 0.0000 & 0.0002  & 0.0001 \\
    	\emph{Cone} & 0.0100 & 0.0008 & 0.0004  & 0.0001 \\
    	\emph{Symm. Clust.} & 0.1560 & 0.0508 & 0.0061 & 0.0078  \\
    	\emph{Shifted Clust.} & 0.1700 & 0.0516 & 0.0078 & 0.0067  \\
    	\emph{Uneven Clust.} & 0.1960 & 0.0896 & 0.8468 & 0.8491  \\
            \hhline{-|----}
            \emph{Normal} & 0.1960 & 0.1580 & 0.1940 & 0.2260\Tstrut\\
    	\bottomrule
    \end{tabular}
\end{sc}
\end{small}
\end{center}
\vskip -0.1in
\label{tab:knn_ovl_results}
\end{table}

\subsection{Gaussian KL-Divergence (GKL)}
A common method for calculating KL-Divergence is to estimate continuous parameters from the observed distribution and then calculate the closed-form of Equation \ref{eq:KL-div}. Here, $\mu_{p}$ and $\sigma_{p}$ are the mean and covariance matrix of the observed distribution, and $d$ is the number of dimensions. We propose a measure based on the closed-form KL-divergence between two multivariate Gaussians in Equation~\ref{eq:GKL}, where the $tr()$ function is the sum of the diagonal elements of the matrix \cite{Duchi2016-gq}. Although we don't expect our example distributions to be well-approximated by a normal distribution, this method shows up frequently in machine learning frameworks (e.g. Variational Autoencoders \cite{doersch2016tutorial}, t-SNE \cite{van2008visualizing}, Reinforcement Learning \cite{filippi2010optimism}).
\begin{equation}
GKL = \frac{1}{2}(\mu_{p}^{\top}\mu_{p}+tr(\sigma_{p})-d-\log|\sigma_p|) \label{eq:GKL}
\end{equation}
With this measure, we were surprised to find that Table \ref{kl_gauss_results} indicates that data spread becomes less complete as we move from the three spherical distributions, to the \emph{Cone} distribution, and to the cluster distributions. However, GKL does not produce results that differ greatly from our other measures, and GKL similarly suffers from the weakness of KL-Divergence measures described in Section~\ref{subsec:entropy}, in that the range for high dimensional distributions is much larger than it is for lower dimensional distributions.

\begin{table}
\caption{Example distribution results for Gaussian KL-Divergence (GKL).}
\vskip 0.15in
\begin{center}
\begin{small}
\begin{sc} 
    \begin{tabular}{l|rrrr}
    	\toprule
    	Example Dist. &
    	\multicolumn{1}{c}{2D}  & \multicolumn{1}{c}{10D} & \multicolumn{1}{c}{50D}     & \multicolumn{1}{c}{100D}\Bstrut\\
    	\hhline{-|----}
    	\emph{Shell} & 0.0013 & 0.0078 & 0.0478 & 0.0985\Tstrut\\
    	\emph{Nested Shell} & 0.0031 & 0.0122 & 0.0662 & 0.1344  \\
    	\emph{Sphere} & 0.0011 & 0.0086 & 0.0500 & 0.0945 \\
    	\emph{Cone} & 0.0926 & 0.7479 & 1.4945 & 1.8939 \\
    	\emph{Symm. Clust.} & 0.5064 & 3.5500 & 18.5039 & 18.4996  \\
    	\emph{Shifted Clust.} & 0.5902 & 4.3510 & 18.5038 & 18.4996  \\
    	\emph{Uneven Clust.} & 0.5141 & 3.6196 & 18.5039 & 18.4996  \\
            \hhline{-|----}
            \emph{Normal} & 0.0069 & 0.0130 & 0.0516 & 0.0993\Tstrut\\
    	\bottomrule
    \end{tabular}
\end{sc}
\end{small}
\end{center}
\vskip -0.1in
\label{kl_gauss_results}
\end{table}
\clearpage
\onecolumn
\section{Extended Results}\label{app:results}
\begin{table}[h!]
\vskip -0.1in
\caption{Example distribution results on Average Cosine Similarity (ACS) and I(V).}
\vskip 0.15in
\begin{center}
\begin{small}
\begin{sc}
	\begin{tabular}{l|cc|cc|cc|cc}
		\toprule
		 Example & \multicolumn{2}{c|}{2D} & \multicolumn{2}{c|}{10D} & \multicolumn{2}{c|}{50D} & \multicolumn{2}{c}{100D}           \\
		Distribution & ACS     & I(V)    & ACS     & I(V)    & ACS   & I(V)    & ACS     & I(V)\Bstrut\\
		\hhline{-|--|--|--|--}
		\emph{Shell} & 0.0027 & 0.9737 & 0.0008 & 0.9916 & 0.0008 & 0.9977 &  0.0007 & 0.9988\Tstrut\\
		\emph{Nested Shell} & 0.0027 & 0.9801 & 0.0008 & 0.9881 & 0.0008 & 0.9964 & 0.0007 & 0.9982 \\
            \emph{Sphere} & 0.0036 & 0.9613 & 0.001 & 0.9905 & 0.0008 & 0.9976 & 0.0007 & 0.9988 \\
		\emph{Cone} & 0.8176 & 0.9469 & 0.5728 & 0.9471 & 0.5201 & 0.9890 & 0.5119 & 0.9944 \\
		\emph{Symm. Clust.} & 0.0020 & 0.7888 & 0.0004 & 0.8857 & 0.0007 & 0.9643 & 0.0007 & 0.9822 \\
		\emph{Shifted Clust.} & 0.0109 & 0.9295 & 0.0108 & 0.8251 & 0.0023 & 0.8360 & 0.0012 & 0.8326 \\
		\emph{Uneven Clust.} & 0.0249 & 0.7797 & 0.0183 & 0.8972 & 0.0076 & 0.9664 & 0.0043 & 0.9819 \\
            \hhline{-|--|--|--|--}
            \emph{Normal} & 0.0022 & 0.9969 & 0.0007 & 0.9934 & 0.0008 & 0.9974 & 0.0008 & 0.9988\Tstrut\\
		\bottomrule
	\end{tabular}
\end{sc}
\end{small}
\end{center}
\vskip -0.1in
\end{table}

\begin{table}[h!]
\vskip -0.1in
\caption{Example distribution results on Eigenvalue Ratio (ER) and Eigenvalue Early Enrichment (EEE).}
\vskip 0.15in
\begin{center}
\begin{small}
\begin{sc}
	\begin{tabular}{l|cc|cc|cc|cc}
		\toprule
		 Example & \multicolumn{2}{c|}{2D} & \multicolumn{2}{c|}{10D}& \multicolumn{2}{c|}{50D} & \multicolumn{2}{c}{100D}           \\
		Distribution &
		ER     & EEE   & ER     & EEE & ER     & EEE   & ER     & EEE\Bstrut\\
		\hhline{-|--|--|--|--}
		\emph{Shell} & 0.9021 & 0.0129 & 0.8380 & 0.0307 & 0.7943 & 0.0354 & 0.7845 & 0.0361\Tstrut\\
		\emph{Nested Shell} & 0.8552 & 0.0195 & 0.7991 & 0.0381 & 0.7550 & 0.0417 &  0.7542 & 0.0422 \\
		\emph{Sphere} & 0.9118 & 0.0115 & 0.8210 & 0.0323 & 0.7845 & 0.0363 & 0.7889 & 0.0353\\
		\emph{Cone} & 0.4173 & 0.1028 & 0.0807 & 0.1155 & 0.0185 & 0.0536 & 0.0090 & 0.0458 \\
		\emph{Symm. Clust.} & 0.1124 & 0.1995 & 0.0126 & 0.4987 & 0.0003 & 0.5503 & 0.0001 & 0.5394 \\
		\emph{Shifted Clust.} & 0.0915 & 0.2081 & 0.0070 & 0.5792 & 0.0021 & 0.6558 &  0.0007 & 0.6746 \\
		\emph{Uneven Clust.} & 0.1102 & 0.2004 & 0.0134 & 0.5067 & 0.0003 & 0.5506 &  0.0001 & 0.5397 \\
            \hhline{-|--|--|--|--}
            \emph{Normal} & 0.7910 & 0.0292 & 0.7917 & 0.0398 & 0.7839 & 0.0368 &  0.7804 & 0.0362\Tstrut\\
		\bottomrule
        \end{tabular}
\end{sc}
\end{small}
\end{center}
\vskip -0.1in
\end{table}

\begin{table}[h!]
\vskip -0.1in
\caption{Example distribution results for Vasicek Ratio MSE (VRM) and Discrete KL-Divergence MSE (DKLM).}
\vskip 0.15in
\begin{center}
\begin{small}
\begin{sc}
	\begin{tabular}{l|rr|rr|rr|rr}
		\toprule
		 Example & \multicolumn{2}{c|}{2D} & \multicolumn{2}{c|}{10D} & \multicolumn{2}{c|}{50D} & \multicolumn{2}{c}{100D}           \\
		Distribution &
		\multicolumn{1}{c}{VRM}     & \multicolumn{1}{c|}{DKLM}  & \multicolumn{1}{c}{VRM}     & \multicolumn{1}{c|}{DKLM}  & \multicolumn{1}{c}{VRM}     & \multicolumn{1}{c|}{DKLM}   & \multicolumn{1}{c}{VRM}     & \multicolumn{1}{c}{DKLM}\Bstrut\\
		\hhline{-|--|--|--|--}
		\emph{Shell} & 0.1891 & 0.1349 & 0.0012 & 0.0003 & 0.0010 & 0.0000 & 0.0010 & 0.0000\Tstrut\\
		\emph{Nested Shell} & 0.0465 & 0.0774 & 0.0013 & 0.0027 & 0.0014 & 0.0025 & 0.0014 & 0.0019 \\
		\emph{Sphere} & 0.0106 & 0.0169 & 0.0011 & 0.0003 & 0.001 & 0.0000 & 0.0009 & 0.0000\\
		\emph{Cone} & 0.0686 & 0.0266 & 0.0603 & 0.0735 & 0.0172 & 0.0521 & 0.0097 & 0.0295 \\
		\emph{Symm. Clust.} & 0.1545 & 0.1434 & 0.1983 & 0.6150 & 0.2655 & 0.7538 & 0.2624 & 0.6200 \\
		\emph{Shifted Clust.} & 0.2143 & 0.5392 & 0.2401 & 0.6748 & 0.2398 & 22.5188 & 0.2386 & 20.5810 \\
		\emph{Uneven Clust.} & 0.1637 & 0.1470 & 0.1990 & 0.8154 & 0.2637 & 0.8012 &  0.2574 & 0.6333 \\
            \hhline{-|--|--|--|--}
            \emph{Normal} & 0.0047 & 0.0190 & 0.0012 & 0.0017 & 0.0010 & 0.0001 &  0.0010 & 0.0000\Tstrut\\
		\bottomrule
	\end{tabular}
\end{sc}
\end{small}
\end{center}
\vskip -0.1in
\end{table}

\begin{table}[h!]
\vskip -0.1in
\caption{Example distribution results on Nearest Neighbor Entropy Ratio (NNR) and Nearest Neighbor KL-Divergence (NNKL).}
\vskip 0.15in
\begin{center}
\begin{small}
\begin{sc}
	\begin{tabular}{l|rr|rr|rr|rr}
		\toprule
		 Example & \multicolumn{2}{c|}{2D} & \multicolumn{2}{c|}{10D} & \multicolumn{2}{c|}{50D} & \multicolumn{2}{c}{100D}           \\
		Distribution &
		 \multicolumn{1}{c}{NNR} & \multicolumn{1}{c|}{NNKL} &  \multicolumn{1}{c}{NNR} & \multicolumn{1}{c|}{NNKL} &  \multicolumn{1}{c}{NNR} & \multicolumn{1}{c|}{NNKL} &  \multicolumn{1}{c}{NNR} & \multicolumn{1}{c}{NNKL}\Bstrut\\
		\hhline{-|--|--|--|--}
		\emph{Shell} & 0.1323 & 7.4573 & 0.9981 & 0.8837 & 1.0028 & 0.3357 & 1.0024 & 0.2532\Tstrut\\
		\emph{Nested Shell} & 0.3996 & 5.2696 & 0.9950 & 1.0005 & 0.9899 & 5.3053 & 0.9887 & 13.2125 \\
		\emph{Sphere} & 0.9749 & 0.9267 & 0.9994 & 0.8082 & 1.0026 & 0.3703 & 1.0023 & 0.2558 \\
		\emph{Cone} & 0.9350 & 1.2659 & 0.9939 & 1.1920 & 1.0022 & 0.4667 & 1.0022 & 0.2853 \\
		\emph{Symm. Clust.} & 0.8971 & 1.4874 & 0.8505 & 14.1981 & 0.7458 & 160.3722 & 0.7135 & 394.5467 \\
		\emph{Shifted Clust.} & 0.8312 & 1.9226 & 0.8111 & 17.5756 & 0.7188 & 177.1980 & 0.6882 & 429.2692 \\
		\emph{Uneven Clust.} & 0.9015 & 1.4215 & 0.8495 & 13.9207 & 0.7457 & 160.4552 & 0.7134 & 394.7038 \\
            \hhline{-|--|--|--|--}
            \emph{Normal} & 0.9973 & 0.7260 & 0.9996 & 0.7083 & 0.9999 & 0.9290 & 1.0001 & 1.1751\Tstrut\\
		\bottomrule
	\end{tabular}
\end{sc}
\end{small}
\end{center}
\vskip -0.1in
\end{table}


\end{document}